\documentclass[11pt]{article}

\usepackage[margin=0.8in]{geometry}

\usepackage{amsfonts, amsmath, amssymb, amsthm}
\usepackage{mathtools} 

\usepackage{bm}
\usepackage{verbatim}
\usepackage{euscript}
\usepackage{graphicx}
\usepackage{paralist}
\usepackage{cancel} 

\usepackage{bbm}

\usepackage{nicefrac}

\usepackage{thmtools,thm-restate}

\newcommand{\abs}[1]{\left \lvert #1 \right \rvert}

\newcommand{\cardinality}[1]{ \left\lvert #1 \right\rvert }
\newcommand{\ceil}[1]{\left \lceil #1 \right \rceil}

\newcommand{\prob}[1]{\mathbb{P} \left\{ #1 \right\}}

\newcommand{\probunder}[2]{\mathbb{P}_{#1} \left\{ #2 \right\}}
\newcommand{\probdistunder}[1]{\mathbb{P}_{#1}}

\newcommand{\expectation}[1]{\mathbb{E} \left[ #1 \right]}

\newcommand{\expund}[2]{\mathbb{E}_{#1} \left[ #2 \right]}
\newcommand{\indicator}[1]{\mathbbm{1} \left\{ #1 \right\}}

\usepackage{enumitem}
\usepackage{graphicx}
\usepackage{soul}
\usepackage{float}

\usepackage{xspace}
\usepackage{wrapfig}

\usepackage[usenames,dvipsnames]{color}
\usepackage{xcolor}

\usepackage[colorinlistoftodos]{todonotes}
\usepackage[colorlinks=true, allcolors=black]{hyperref}

\usepackage{natbib}

\usepackage{algorithmicx, algpseudocode}
\usepackage[ruled]{algorithm}

\usepackage[normalem]{ulem}

\newcommand\numberthis{\addtocounter{equation}{1}\tag{\theequation}}

\newtheorem{lemma}{Lemma}
\newtheorem{corollary}{Corollary}

\newtheorem{definition}{Definition}

\newtheorem{remark}{Remark}

\newtheorem{claim}[lemma]{Claim}

\newenvironment{remark-nolabel}{\noindent{\bf Remark}\hspace*{1em}}{\bigskip}

\newcommand{\muhat}{\widehat{\mu}}

\hypersetup{
  colorlinks   = true, 
  urlcolor     = blue, 
  linkcolor    = blue, 
  citecolor   = blue 
}

\long\def\comment#1{}

\usepackage{caption}
\usepackage{subcaption}

\begin{document}

\newcommand{\AlgGE}{\textsc{SnoozeIt}} 
\newcommand{\Alg}{\textsc{Alg}}
\newcommand{\DelTil}{\widetilde{\Delta}}
\newcommand{\lambTil}{\widehat{\lambda}}
\newcommand{\lambTilGreat}{>_{\lambTil}}
\newcommand{\Arms}{\mathcal{A}}
\newcommand{\GoodEvent}{\mathcal{G}}

\newcommand{\swucb}{\textsc{Sw-Ucb\#}}
\newcommand{\rexp}{\textsc{Rexp3}}
\newcommand{\exps}{\textsc{Exp.S}}

\newcommand{\bern}[1]{Ber \left( #1 \right)}
\newcommand{\kl}[2]{KL \left( #1, #2 \right)}

\newcommand{\ag}[1]{\textcolor{red}{{\bf AG: #1}}}

\title{\bfseries On Slowly-varying Non-stationary Bandits}
\author{Ramakrishnan Krishnamurthy\thanks{Indian Institute of Science. {\tt kramakrishna@iisc.ac.in}} \quad Aditya Gopalan\thanks{Indian Institute of Science. {\tt aditya@iisc.ac.in}}}

\date{}
\maketitle

\begin{abstract}

We consider minimisation of dynamic regret in non-stationary bandits
with a slowly varying property. 
Namely, we assume that arms' rewards are stochastic and independent over time,
but that the absolute difference between the expected rewards
of any arm at any two consecutive time-steps is at most a drift limit $\delta > 0$.
For this setting that has not received enough attention in the past, 
we give a new algorithm which extends naturally the well-known Successive
Elimination 
algorithm to the non-stationary bandit setting.
We establish the first instance-dependent regret upper bound
for slowly varying non-stationary bandits. 
The analysis in turn relies on a novel characterization of the instance 
as a {\em detectable gap} profile that depends on the expected arm reward differences.
We also provide the first minimax regret lower bound for this problem,
enabling us to show that our algorithm is essentially minimax optimal.
Also, this lower bound we obtain matches that of the more
general total variation-budgeted bandits problem, 
establishing that the seemingly easier former problem is at least as hard as the
more general latter problem in the minimax sense. 
We complement our theoretical results with experimental illustrations.

\end{abstract}
\section{INTRODUCTION}

Reinforcement learning, and specifically bandit optimization, in dynamically changing environments has remained an active topic of study in machine learning. A variety of non-stationary bandit settings have been studied incorporating a range of structural assumptions. At one end are classical stochastic models such as restless bandits \citep{whittle1988restless}, where the state of the arms governs the bandit problem at any instant, but the transitions between these problems (states) follow probabilistic dynamics. At the other extreme are settings with non-stochastic and arbitrarily changing rewards such as prediction with expert advice (and the EXP3 algorithm)\citep{cesa2006prediction, auer2002nonstochastic}.
In between these extremes lie settings of changing environments where the adversary (environment) is assumed to be limited in its ability to change the rewards, i.e., a structural constraint is put on the amount of change in the rewards across time. These include the abrupt change (or switching experts) model \citep{garivier2011upper}, where at most $k$ arbitrary changes to the reward distributions are allowed in the entire time horizon, and the variation-budgeted (drifting) change model \citep{besbes2014stochastic}, in which the total magnitude of changes (of rewards) across successive time steps is constrained to be within an overall budget.

In this paper, we focus on slowly-varying bandits -- a different and arguably commonly encountered, yet less studied, model of non-stationary bandits. 
In this setting, the arms are allowed to change arbitrarily over time as long as the amount of change in their mean rewards between two successive time steps is bounded uniformly across the horizon.
Many real-world settings naturally involve observables whose distributions are `smooth' over time, in the sense that their instantaneous rate of change is not too large, e.g., slowly drifting distributions in natural language tasks \citep{lu2020countering}, data from physical transducers (position, velocity, power, temperature, chemical concentration), and slowly fading wireless channels \citep{tse2005fundamentals}. Though the slowly-varying bandit setting is subsumed by the total variation-budgeted model with an appropriate budget, the hope is that the local smoothness of the rewards with time can be exploited by the learner to perform better. We make the following contributions in this regard. 

\begin{enumerate}
	\item We give a new algorithm for slowly-varying two-armed bandits. The algorithm is based on the principle of adaptive exploration followed by commitment in phases -- a strategy that is known to be optimal for stationary stochastic bandits -- but where the commitment phase is adjusted depending on the smoothness constraint on the rewards. The design of the algorithm also involves the estimation of a novel property of the bandit instance called the detectable gap profile, which essentially characterises the gap between the local averaged means of the arms that can be reliably detected, across time.
	
	\item We derive a regret bound for this algorithm in terms of the detectable gap profile of the bandit instance (Theorem \ref{thm:inst-dep-reg}). The bound, to our knowledge, is the first instance-dependent regret guarantee for any algorithm for drifting stochastic bandits. Moreover, the worst case regret bound for a horizon $T$, over all instances for a given constraint $\delta$ on instantaneous reward change, is shown to be $O(T \delta^{\nicefrac{1}{3}})$ (Theorem \ref{thm:minimax-bound}).
	
	\item We complement this worst case regret upper bound with a matching fundamental lower bound (Theorem \ref{thm:lower-bound}) -- the first of its kind for slowly-varying bandits -- that shows that our algorithm is order-wise minimax-optimal. Interestingly, the minimax regret rate happens to be the same as that of the total variation-budgeted (with equivalent budget $\delta T$) setting, establishing that the more constrained slowly-varying setting is at least as hard (in a worst case sense) as the more general total variation-budgeted setting. 
	
	\item We numerically evaluate the performance of our new algorithm and existing approaches, showing that there are situations in which it is able to outperform the other approaches. 
\end{enumerate}

\subsection{Related Work}
\label{subsec:related-work}

Non-stationarity in bandit optimization has been extensively studied and dates back to the seminal work of \cite{whittle1988restless} on restless bandits, in which arms' states (and thus their reward distributions) change according to Markovian dynamics (also see \cite{slivkins2008adapting}). The past few decades, however, have witnessed the growth of hybrid adversarial and stochastic bandit models for non-stationarity, where the distributions of arms across time can be set by an adversary ahead of time in an arbitrary manner. Among the many  examples of such models are the following; the abruptly changing (or switching) bandits setting \citep{garivier2011upper, auer2019adaptively}, where the distributions are piecewise stationary and can arbitrarily change at unknown time steps,
and then, the the total variation-budgeted setting or the drifting bandits setting introduced by \cite{besbes2014stochastic, besbes2019optimal}. The authors show that with a constraint on the total amount of successive changes in arms' reward means, it is possible to devise algorithms with guarantees on the dynamic regret that depend on the variation budget $V_T$ and total time horizon $T$ of order roughly $O(V_T^{\nicefrac{1}{3}} T^{\nicefrac{2}{3}})$ (ignoring logarithmic factors).

In this drifting bandits regime, there have subsequently been multiple works 
that have focused on dropping assumption of knowledge of drift parameter \citep{karnin2016multi},
extension of that to contextual bandits \citep{luo2018efficient},
and achieving optimal minimax bounds therein \citep{chen2019new},
however, an instance-dependent regret bound characterization has so far remained elusive under any sort of drifting reward constraint.
Our results not only match these optimal regret rates in a minimax sense, but we also give a more instance-dependent regret characterisation for our algorithm in a slowly-varying drifting bandits setting. 

In terms of lower bounds, \cite{besbes2014stochastic} provide a lower bound of $\Omega\left(V_T^{\nicefrac{1}{3}} T^{\nicefrac{2}{3}} \right)$ that matched their algorithm's upper bound (upto logarithmic factors), 
thereby closing the door on any possible improvements to them in recent years.
However, we show a matching minimax lower bound for slowly-varying bandits with equivalent per-round drift limit, which is a stronger result as it applies to a more restricted class of problems instances.

Perhaps the only other work that studies the exact slowly-varying non-stationary bandit setting that we consider here is that of \cite{wei2018abruptly}. They modify the sliding-window-UCB algorithm of \cite{garivier2011upper} to employ windows that grow in size with time to get the SW-UCB\# algorithm. 


\section{SETTING AND PRELIMINARIES}

We consider bandits with arm set $\Arms = \{1,2\}$
\footnote{We restrict to two arms because we would like to focus on understanding the dependence of regret on the non-stationarity over \emph{time} as opposed to over the arm space. 
This is also the line of investigation adopted in other works on non-stationary bandits for similar reasons 
(\cite{auer2018adaptively}, \cite{karnin2016multi})} and a time horizon of $T$.
At time $t \in [T]:=\{1,2,\dots,T \}$, when arm $a \in \Arms$ is pulled by a bandit algorithm, depending on only past history, a stochastic reward is drawn from a Bernoulli
\footnote{In general, our theoretical analysis and results hold for any sub-gaussian reward distributions with a suitably bounded variance.} distribution with expected value $\mu_{a,t} \in [0,1]$ is obtained.
We denote by $\mu_t := (\mu_{1,t}, \mu_{2,t})$ the expected reward tuple at time $t \in [T]$, 
and denote by $\mu^*_t = \max_{a \in \Arms} \mu_{a,t}$ the optimal arm (in terms of expected reward) at time $t$.
Write $\mu := (\mu_1, \mu_2, \dots, \mu_T)$ be the expected reward profile of the bandit instance.

A bandit instance $\mu$ is defined to be  \emph{slowly-varying} with \emph{drift limit} $\delta > 0$ (denoted as $\mu \in S_\delta$) if the expected reward profile satisfies
\begin{align*}
    \forall a \in \Arms, t \in [T-1], \abs{\mu_{a,t}-\mu_{a,t+1}} \leq \delta. \numberthis \label{eqn:drift-limit}
\end{align*}
In other words, for an arm $a \in \{1, 2\}$ at a time step $t$, the expected reward $\mu_{a,t}$ can drift in value to some $\mu_{a,t+1}$ (the next time-step) by at most $\delta$.




\begin{definition}[Regret]
    For a bandit instance $\mu$, the (expected) regret  
    incurred by an algorithm $\Alg$ is
    \begin{equation*}
        R(\Alg) = \sum_{t=1}^T \mu^*_t - \expectation{\mu_{\Alg(t), t}},
    \end{equation*}
    where $\mu^*_t = \max_{a \in \Arms} \mu_{a,t}$ is the mean reward of the optimal arm at time $t$,
    and $\Alg(t) \in \Arms$ is the arm pulled by $\Alg$ at time $t \in [T]$.
\end{definition}

Note that this is the \emph{dynamic regret}, where 
the performance benchmark at each time-step is the expected reward of the optimal arm at that time-step ($\mu^*_t$).
This is a stronger (i.e., harsher) notion of regret compared to the classical notion of \emph{static regret}, where
the benchmark at all time-steps is the expected reward of the best single arm across the entire horizon in hindsight.

The goal is to learn to play arms to achieve low (dynamic) regret
for any problem instance $\mu \in  S_\delta$.
 
We assume that the algorithm has access to the drift limit $\delta$ (or a suitable upper bound on it). 
This is reasonable as in practice
there is often domain specific information available in advance about the drift of the 
quantity in consideration.

We now introduce a novel characterization of a non-stationary bandit instance, and highlight its significance to the algorithm design and analysis to follow later.


\paragraph{Gaps and Detectable Gap Profile.}

\begin{figure*}[!h]
    \centering
    \makebox[\textwidth][c]{
    \includegraphics[width=1.1\linewidth]{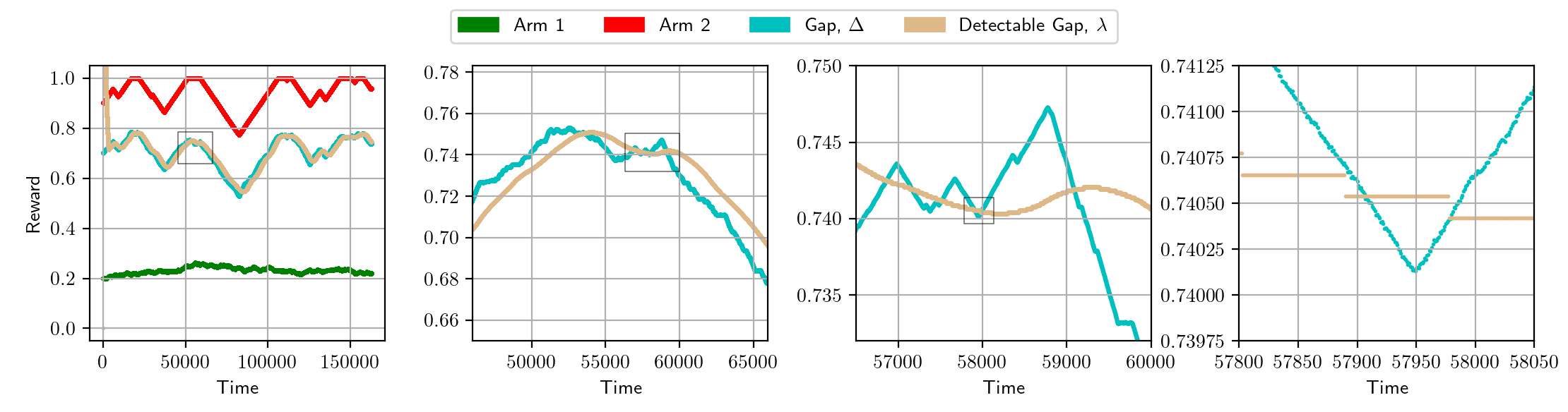}
    }
    \caption{Depiction of the arms' true reward means, the gap profile, and the detectable gap profile of one problem instance illustrated at 4 zoom levels. The box in each image is the region zoomed in for the next image (to its right).
    First, the detectable gaps ($\lambda$) reasonably positively correlates with the gaps ($\Delta$), and initially (in general, whenever there is not a sufficient window size to detect a gap), for a short span, it takes the form $\sqrt{\frac{c_0 \log T}{t}}$.
    Second, $\lambda$ roughly trails $\Delta$ as it depends on a recent window of $\Delta$ at every time-step.
    Third, $\lambda$ is `smoother' compared to $\Delta$, as by design, its values are averaged over a window of samples.
    Fourth, although $\lambda$ is described as a continuous optimization problem over $[0,1]$, 
    it actually transforms into a discrete optimization problem over different integral window sizes $w$. Thus, $\lambda$ has a `piece-wise constant' appearance.}
    \label{fig:lambda-description}
\end{figure*}

In the {\em stationary} stochastic (two-armed) bandit problem, the suboptimality \emph{gap} $\Delta := \abs{\mu_1 - \mu_2}$ essentially characterises the attainable regret rate\footnote{assuming bounded/sub-gaussian rewards}. 
With {\em non-stationary} reward distributions, the notion of a (time-invariant) gap must be generalised to a {\em gap profile}
$\Delta := (\Delta_t)_t$ where $\Delta_t := \abs{\mu_{1,t} - \mu_{2,t}}$ is the difference 
at time $t$ between the expected rewards of the two arms, $\mu_{1,t}$ and $\mu_{2,t}$.



We now introduce the notion of a {\em detectable gap profile} $\lambda := (\lambda_t)_t$, which intuitively helps to characterise how hard it is to reliably estimate which arm is optimal and by how much at any time $t$. This is a derived quantity expressed in terms of the gap profile taken over a local window of time leading up to $t$. %
Precisely,
we define 
\begin{align*}
    \lambda_t = 
    \begin{dcases}
    	\max \left\{ \lambda \in [0,1] : \frac{1}{w(\lambda)}\abs{\sum_{t'=s}^t \mu_{1,t'} - \mu_{2,t'}} \geq \lambda  \right\}
        \quad &\text{, if such a $\lambda \in [0,1]$ exists.} \\ 
    	\sqrt{\frac{c_0 \log T}{t}} \quad &\text{, otherwise.}
    \end{dcases}
    \numberthis \label{eqn:lamdba-curve}
\end{align*}

Here, $s:=t-w(\lambda)+1$ stands for the starting time-step of a sampling window
of size $w(\lambda) := \ceil{\frac{c_0 \log T}{\lambda^2}}$ that terminates at $t$, and $c_0 = 144$ is a constant \footnote{In fact, in the remainder of the paper, all notations of the form $c_i$ shall be suitable universal constants.}. Intuitively, $\lambda_t = \lambda$ for some time-step $t$ implies that a gap of $\lambda$ can be detected with high probability by observing samples from both arms in the past $\approx \nicefrac{1}{\lambda^2}$ time-steps.
Fig. \ref{fig:lambda-description} depicts the detectable gap profile and gap profile for a two-armed bandit instance. 

We shall see that the detectable gap profile better reflects the nature of a non-stationary problem instance. 
Specifically, for our algorithm, we shall obtain a regret upper bound (to be shown in Theorem \ref{thm:inst-dep-reg}) 
as a function of the detectable gap profile, $\lambda$, which is an instance-dependent quantity.

\section{ALGORITHM DESCRIPTION}
\label{sec:algorithm}

Successive Elimination (SE) is a well-known algorithmic recipe for solving stationary stochastic bandits, built upon the Explore-Then-Commit (ETC) paradigm 
(see \cite{slivkins2019introduction} for a text-book treatment of SE and ETC)
In its classical form, SE adaptively pulls both arms alternately until it distinguishes the optimal arm from the sub-optimal (in terms of reward means) with a high probability. Then, it drops the inferior arm and indefinitely plays the superior arm.
Our algorithm, $\AlgGE$ (Algorithm \ref{alg:gauge-enjoy}), is a graceful adaptation of the Successive Elimination paradigm to the non-stationary bandits setup.
It pulls arms alternately until it can assertively distinguish the two arms in a `local average' sense, 
and then plays only the optimal arm for a period of time until the sub-optimal arm, due to the non-stationarity, can become the optimal arm.

Specifically, the $\AlgGE$ algorithm runs in a series of episodes.
Every episode begins with an \emph{active phase}, where $\AlgGE$ pulls the two arms alternately (in a round-robin fashion). 
At the end of every time-step, it performs a statistical test (Line \ref{line:stat-test} ) to detect a \emph{clear gap} (in reward means) between the two arms.
If the test succeeds, the active phase ends and the observed sub-optimal arm is possibly
\emph{snoozed} for a certain period of inactivity 
(Lines \ref{line:check-snooze-condition}-\ref{line:snooze}),
termed as the \emph{passive phase}.
During this passive phase, $\AlgGE$ plays only the observed optimal arm.
At the end of this passive phase, the observed sub-optimal arm \emph{respawns} to become active, and the next episode begins.

We introduce some notations to describe the statistical test and for the analysis in the remainder of the paper.

Let $e$ be the total number of episodes (indexed as $1,2,\dots,e$) in a run of $\AlgGE$.
For every episode $i \in [e]$, let $t_i$ denote the time after which the active phase of episode $i$ begins. Note that $t_{i+1}$ corresponds to the time at which episode $i$ ends. We also always have $t_1 = 0$ and $t_{e+1} = T$, the end time of the final episode.
Let $g_i$ denote the time at which the statistical test on line \ref{line:stat-test} of the algorithm passes, 
and write $\tau_i := g_i - t_i$ to denote the duration of the active phase of episode $i$. %

Write $\muhat_{a,t}(w)$ to denote the empirical reward mean (or simply empirical mean) of arm $a$
at time $t$ measured/calculated from it's last $w$ pulls, i.e., the most recent $w$ pulls until time $t$.
Note that the true reward means $\mu_{a,t'}$ for different time-steps $t'$
corresponding to these last $w$ pulls 
need not be the same.

\begin{definition}[$\lambTil$-betterness]
\label{def:lambda-better}
Let $\lambTil > 0$. At time $t$, in the active phase of episode $i$,
arm $a$ is said to be $\lambTil$-better than arm $b$ 
(written as $a \lambTilGreat b$) if, 
for a window sample count $w := \ceil{\frac{c_1 \log T}{\lambTil^2}} \leq \frac{t-t_i}{2}$, 
we have $ LCB_{a,t}(w) > UCB_{b,t}(w) + 2 r(w) - \delta$,
and the inequality holds for no other $\lambTil' < \lambTil$.
\end{definition}

In the above definition, write
$LCB_{a,t}(w) := \muhat_{a,t}(w) - r(w)$  
($UCB_{b,t}(w) := \muhat_{b,t}(w) + r(w)$)
to denote the lower (upper) confidence bound of arm $a$'s ($b$'s) recent reward mean,
and $r(w) := \sqrt{\frac{2\log T}{w}}$ 
is the accuracy radius with $w$ samples.
The constraint $w \leq \frac{t-t_i}{2}$ ensures that
the samples used in the statistical test (or comparison expression in the definition) are all from the (current) episode $i$.

In the statistical test (Line \ref{line:stat-test}) at a time $t$, 
$\AlgGE$ evaluates if some arm is $\lambTil$-better than the other (as in definition \ref{def:lambda-better})

Essentially, it compares the empirical means of the two arms
measured over a window of samples ending at $t$, the $\muhat_{a,t}(w)$s. 

The size of the window, $w$, in which the empirical means are calculated is dynamically chosen based on the empirical detectable gap, $\lambTil$, istelf that is being tested for.
If the test passes, then $\AlgGE$ computes 
a sub-optimality buffer period of duration $\text{buf}=\frac{2}{\delta} \sqrt{\frac{\log T}{\tau_i}}$ (Line \ref{line:compute-buffer}), 
that began no earlier than the start of the current episode $t_i$.
By time $t$, if the buffer period has not fully elapsed, 
we snooze the sub-optimal arm for a passive phase that runs until the end of the buffer period.

\begin{remark}
\label{rem:computation-in-stat-test}
    The statistical test (Line \ref{line:stat-test}) describes the identification of a $\lambTil$-better arm 
    as an optimization problem over the continuous domain $\lambTil \in [0,1]$. 
    However, with the integrality constraint of the sample window size $w$ corresponding to a $\lambTil \in [0,1]$,
    we have $w \in D \subseteq \{1,2, \dots, t\}$ belongs to a finite countable domain of size not more than $t$.
    Thus, the identification of a $\lambTil$-better arm becomes a discrete optimization problem,
    and indeed, the statistical test is deterministic and tractable. 
\end{remark}

\begin{algorithm}[h]
\small
\caption{$\AlgGE$: Plays the non-stationary slowly-varying bandit problem instance}
\label{alg:gauge-enjoy}
{\bf Input:} Time horizon $T$, a set of two arms $\Arms=\{1,2\}$ with sample access, the drift limit $\delta$.\\
{\bf Output:} Play an arm for every time-step. 
\begin{algorithmic}[1]
    \State Initialize set of active arms $A = \Arms$,
    set of snoozed arms $S = \emptyset$.
    \State Initialize episode index $i \leftarrow 1$, $\tau_1 = 0$.

    \For{$t = 1,2,\dots,T$}
        \State $x \leftarrow$ Least recently pulled arm in $A$.
        \Comment{Play active arms in round-robin fashion}
        \State Pull arm $x$ and observe reward $\muhat_{x,t}$.
        \label{line:pull}
        \If{$\exists$ arms $a,b \in A$, $\exists \lambTil \in [0,1]$ s.t. $a \lambTilGreat b$}
        \label{line:stat-test}
        \Comment{As in Definition \ref{def:lambda-better}}
            \State Statistical test success time, $g_i = t$
            \State Active phase duration, $\tau_i = g_i - t_i$.
            \State Sub-optimality buffer, $\text{buf} = \dfrac{2}{\delta} \sqrt{\dfrac{\log T}{\tau_i}}$
            \label{line:compute-buffer}
            \If{$\text{buf} > \tau_i$}
            \label{line:check-snooze-condition}
                \State $A \leftarrow A \setminus \{b\}$
                \State $S \leftarrow S \cup \{(b, t_i + \text{buf})\}$
                \label{line:snooze}
                \Comment{Snooze arm}
            \Else
                \State $i \leftarrow i+1, t_i \leftarrow t$
                \Comment{End episode without passive period}
            \EndIf
        \EndIf
 
        \If{$\exists (x,s) \in S: s \geq t$}
            \State $S \leftarrow S \setminus \{(x,s)\}, 
                     A \leftarrow A \cup \{x\}$
            \Comment{Respawn arm}
            \State $i \leftarrow i+1, t_i \leftarrow t$
            \Comment{End episode for passive period elapses}
        \EndIf
    \EndFor
\end{algorithmic}
\end{algorithm}

Optionally, we refer the reader to appendix \ref{appn-subsec:missing-figures} (figures \ref{fig:alg-trajectory-1} and \ref{fig:alg-trajectory-2}) for a graphical example of an algorithmic trajectory
that illustrates the active phases, statistical tests, and snooze periods.

\section{THEORETICAL GUARANTEES}
\label{sec:guarantees}

Our first main result is a regret bound for the $\AlgGE$ algorithm in terms of the detectable gap profile of a slowly-varying bandit instance. 

\begin{restatable}{theorem}{TheoremInstanceDependentUpperBound}[Instance-dependent regret bound]
\label{thm:inst-dep-reg}
    If $\AlgGE$ is run with a drift limit parameter $\delta$ on a problem instance $\mu \in S_\delta$,
    then its expected regret satisfies
    \begin{align*}
         R(\AlgGE) \leq c_8 + c_6 \sum_{j=1}^{m} \frac{1}{\lambda_{min}(j)}. \log T + c_7,
    \end{align*}
    where $m = \nicefrac{T}{\tau}$ is the number of blocks, 
    each of length not more than $\tau = \min\left\{T, c_3 \delta^{\nicefrac{-2}{3}} \log^{\nicefrac{1}{3}} T\right\}$,
    and for every block $j \in [m]$ spanning a time period $b_j := [(j-1)\tau + 1, j\tau] \cap [T]$, 
    $\lambda_{min}(j) := \min_{t \in b_j}\lambda_t$.
    Here, $c_i$s are suitable constants.
\end{restatable}

We also have the following upper bound on the worst-case (over instances in $S_\delta$) regret for \AlgGE.

\begin{restatable}{theorem}{TheoremMinimaxUpperBound}[Minimax Upper bound]
\label{thm:minimax-bound}
    If $\AlgGE$ is run with a drift limit parameter $\delta$ on a problem instance $\mu \in S_\delta$, $\AlgGE$ incurs an expected regret of
    $O\left(T \delta^{\nicefrac{1}{3}} \log^{\nicefrac{1}{3}} T \right)$.
\end{restatable}

We finally complement the worst-case regret upper bound for $\AlgGE$ with a matching universal minimax regret lower bound: 

\begin{restatable}{theorem}{TheoremLowerBound}[Minimax Lower Bound]
\label{thm:lower-bound}
For any algorithm $\Alg$ and a drift limit $\delta > 0$, there exists a problem instance $\mu \in S_\delta$ such that,
$\Alg$ incurs a expected regret of
$\Omega\left(T \delta^{\nicefrac{1}{3}}\right)$.
\end{restatable}



\paragraph{Discussion} The minimax regret lower bound we obtain establishes (constructively) that if the drift limit $\delta = \Omega(1)$, then it is impossible to achieve a sub-linear (in time) regret for any algorithm.
The interesting problem space is thus when $\delta = o(1)$\footnote{as a function of total time $T$}.

A basic sanity check is to evaluate our results for the stationary bandits setting, that is when $\delta = 0$, and the gap $\Delta$ is unchanged over time.
In that case, in Theorem \ref{thm:inst-dep-reg}, the size of a block is $\tau=T$.
From the detectable gap definition, either with $\lambda_t = \Delta$ from the first assignment, or $\lambda_t > \Delta$ from the second,
we have $\lambda_t \geq \Delta$ at all times $t$.
This gives us a regret bound of $O\left( \frac{1}{\Delta} \log T\right)$. One can also observe that $\AlgGE$ behaves similar to the classical Successive Elimination algorithm in this regime.
After it distinguishes the optimal arm from the sub-optimal, that is, the statistical test passes,
it computes a sub-optimality buffer $\text{buf}=\frac{2}{\delta} \sqrt{\frac{\log T}{\tau_i}} = \infty$, an infinite passive phase,
i.e., it snoozes the sub-optimal arm indefinitely.

Moving on, for non-stationary instances with small positive drift limits -- specifically, 
for $\delta \leq O\left(\frac{\log^{\nicefrac{1}{2}}T}{T^{\nicefrac{3}{2}}} \right)$ -- applying 
Theorem $\ref{thm:inst-dep-reg}$ still yields a block size of $\tau=T$. With a similar analysis, this results in a regret bound of $O\left( \frac{1}{\Delta} \log T\right)$\footnote{With the drift, consider the smallest value of $\Delta_t$ for the expression, or corresponding $\lambda_t$ value.}. 
For $\Delta = \Theta(1)$, it is noteworthy that our results show a logarithmic regret despite the mild non-stationarity of the instance, which, to the best of our knowledge, has not been shown before.

This result in Theorem \ref{thm:minimax-bound} is comparable with that of \cite{besbes2014stochastic} who work on a total variation-budgeted setting.
In our setting, a drift limit of $\delta$ per time-step translates to a cumulative drift limit of $T \delta$ over the entire time horizon.
Precisely, for an arm $a$,
$\sum_{t=1}^{T-1} \abs{\mu_{a,t} - \mu_{a,t+1}} \leq T \delta$.
This cumulative drift limit quantity is termed the variation budget $V_T$ in theirs,
and we, here, have $V_T = T \delta$.
Substituting this in their regret upper bound $O\left(T^{\nicefrac{2}{3}} V_T^{\nicefrac{1}{3}} \log ^{\nicefrac{1}{3}} T\right)$,
we get $O\left(T \delta^{\nicefrac{1}{3}} \log^{\nicefrac{1}{3}} T \right)$, the same upper bound as ours.
Thus, we note that our minimax upper bound matches that of \cite{besbes2014stochastic} which accomodates a more general setting.
On a more intereting note,
substituting $V_T = T \delta$ in their regret lower bound $\Omega(T^{\nicefrac{2}{3}} V_T^{\nicefrac{1}{3}})$,
we get $\Omega\left(T \delta^{\nicefrac{1}{3}} \right)$, the same lower bound expression that we have established.
Thus, we note that our minimax lower bound matches that of \cite{besbes2014stochastic},
and crucially establishes that the seemingly easier slowly-varying bandits problem 
is at least as hard as the more general budgeted drifting bandit problem in a minimax sense.

\section{PROOF SKETCHES}

The formal proofs of all results are deferred to the appendix (\ref{appn-sec:instance-upper-bound}, \ref{appn-sec:minimax-upper}, and \ref{appn-sec:low-bound}) in the interest of space. 


\paragraph{Proof sketch for Theorem \ref{thm:inst-dep-reg} (Instance-dependent regret bound)} 
We establish useful properties about
the nature of the active and passive phases, the sizes of episodes,
the connection between the regret and the detectable gap, $\lambda$.

We first restrict our attention to a \emph{good} event, $\GoodEvent$, where all empirical means fall close to the reward means, 
which occurs with high probability. 

Then (Lemma \ref{lem:inactive-not-optimal} in the appendix), we show that
in every episode, a snoozed arm (if it exists) remains sub-optimal for the entirety of the passive phase.
Towards this, we show that,
for an episode $i$, when the statistical test passes, at some point of time (say $t'$) in the active phase,
we have the true reward means of the arms well separated, i.e., $\mu_{a,t'} - \mu_{b,t'} \geq f$ for some gap $f$.
Then, we argue that the snooze period is carefully chosen based on this guaranteed true gap $f$,
the drift limit $\delta$, and the active phase duration $\tau_i$ such that arm $b$ remains sub-optimal in it. With this, we note that only an optimal arm is played in the passive phase, and thus, 
our algorithm incurs regret only during the active phase, which we try to bound next.

Next, we upper bound the regret $R_i$ in episode $i$ based on $\tau_i$, the duration of the active phase.
We use the fact that the statistical test did not pass at time $g_i - 1 = t_i + \tau_i - 1$,
specifically the fact that no arm was $\lambTil$-better than the other. 
We deduce that, their empirical means, and by the good event assumption, also their true means (over a period of time) were not well separatable after alternately playing arms for $\tau-1$ time steps.
Put simply, for these $\tau-1$ time-steps, the gap between the arms was less. 
This helps us get the following bound for $R_i$, the regret in episode $i$ as a function of the active phase duration $\tau_i$: 
\begin{align*}
        R_i &\leq  1 + 6 \sqrt{(\tau_i-1) \log T}. \numberthis \label{sketch:eqn-1}
\end{align*}

This expression is also used (as a starting step) in the minimax regret upper bound derivation.
However, here, we go on to make a connection between $\tau_i-1$ and the detectable gap profile, $\lambda$.

Based on the design of detectable gap, we show, in Lemma \ref{lem:gauge-lambda-curve-relation}, that, if, for some $\tau >0$, at time $t'=t_i + \tau$ we have
$\lambda_{t'} > \sqrt{\frac{c_2 \log T}{\tau}}$, then our algorithm will surely (under occurrence of $\GoodEvent$) detects
a gap sufficient to declare $a >_{\lambTil} b$, and the statistical test shall pass. However, given that the statistical test did not pass at $g_i - 1 = t_i + \tau_i - 1$, 
we derive the instance-dependent episodic regret bound $R_i \leq c_4. \frac{1}{\lambda_{g_i - 1}} \log T + c_5$.


We next collect all the episodic regrets to obtain the total regret,
For this, we analyze the episode durations in two possible cases:
episodes with a passive phase, and episodes without a passive phase, and show (Lemma \ref{lem:size-of-episode}) that in both cases, 
the minimum duration of any episode is $x = c_3 \delta^{\nicefrac{-2}{3}} \log ^{\nicefrac{1}{3}} T$. We also partition the time horizon $[T]$ into blocks of size $x$, and argue that each block can intersect with at most two episodes
as no episode can be fully contained within a block.

Finally, we account the regret of an episode to a block it intersects with, 
and aggregate over blocks the instance dependent regrets of episodes, via the per-episode regret bound 
to arrive at the final instance-dependent regret bound.


\paragraph{\bf Proof sketch of Theorem \ref{thm:minimax-bound} (Minimax regret bound)} 

For every episode $i$, we upper bound the \emph{average regret} or per time-step regret defined as $R^{ave}_i = O\left(\frac{R_i}{ t_{i+1}-t_i}\right)$.
Then, we shall argue that the regret is bounded by $T$ times the maximum average regret incurred by any episode.

From eqn. \ref{sketch:eqn-1}, we have the regret in an episode $i$ dependent on the active phase duration $\tau_i$ 
as $R_i = O\left( \sqrt{\tau_i \log T} \right)$.

We analyse $R^{ave}_i$ based on the following two cases based on the presence or absence of a passive phase.

\paragraph{Case $\tau_i < \text{buf}$ :}  
Equivalently, we have $\tau_i < \frac{2}{\delta} \sqrt{\frac{\log T}{\tau_i}}$, and thus $\tau_i < 2^{\nicefrac{2}{3}} \delta^{\nicefrac{-2}{3}} \log^{\nicefrac{1}{3}} T$.
As this episode has a passive phase, the duration of this episode, $t_{i+1} - t_i = \text{buf} = \frac{2}{\delta} \sqrt{\frac{\log T}{\tau_i}}$. 
With these, the average regret is bounded as $R^{ave}_i = O\left(\dfrac{\sqrt{\tau_i \log T}}{\frac{1}{\delta}\sqrt{\frac{\log T}{\tau_i}}}\right) = O\left(\tau_i \delta\right) = O\left( \delta^{\nicefrac{1}{3}} \log^{\nicefrac{1}{3}} T \right)$.

\paragraph{Case $\tau_i \geq \text{buf}$ :} Equivalently, $\tau_i \geq \frac{2}{\delta} \sqrt{\frac{\log T}{\tau_i}}$ and thus,
$\tau_i \geq 2^{\nicefrac{2}{3}} \delta^{\nicefrac{-2}{3}} \log^{\nicefrac{1}{3}} T$.
Here, the episode $i$ does not have a passive phase. Thus, the duration of this episode is $t_{i+1} - t_i = \tau_i$.
With these, the average regret is bounded as 
$ R^{ave}_i = O \left( \frac{ \sqrt{\tau_i \log T} }{\tau_i} \right) 
= O\left(\sqrt{\frac{\log T}{\tau_i}} \right) 
= O\left(\delta^{\nicefrac{1}{3}} \log^{\nicefrac{1}{3}} T \right) $.

From both cases, as the maximum per-time step regret is similarly bounded,
we multiply this by the length of the horizon, $T$, to show the desired minimax bound of $O\left(T \delta^{\nicefrac{1}{3}} \log^{\nicefrac{1}{3}} T\right)$.

Optionally, we refer the reader to appendix \ref{appn-sub-sec:ave-reg-by-epis-variables} for 
an assessment of the interplay between average regret in an episode, its total duration, and its active phase duration.



\paragraph{Proof sketch for Theorem \ref{thm:lower-bound} (Worst-case regret lower bound)}

We derive this result with information-theoretic arguments commonly employed in bandit literature \cite{garivier2016optimal}.

First, we adapt the standard change of measure inequality for sequential sampling (see e.g., \cite{garivier2019explore}) to a version that can accommodate non-stationary reward distributions.
Second, we break the time horizon into suitably small blocks of size $m$, establish lower bounds within each of them, 
and finally aggregate them to arrive at the lower bound expression.
Towards proving the lower bound within a block, we consider a base instance $\nu$, where both arms are identical and stationary with $\bern{\nicefrac{1}{2}}$ reward distributions.
We design a confusing instance, $\nu'$, where one arm is stationary with $\bern{\nicefrac{1}{2}}$ rewards, whereas,
the other arm (say, arm $1$, with a distribution of $\nu'_{1,t}$ at time $t$ within the block) exhibits non-stationarity as follows: $ \nu'_{1,t}$ is a Bernoulli distribution with parameter $\frac{1}{2} + \frac{t-1}{m} \varepsilon$ if $t \leq \ceil{\frac{m}{2}}$, and $\frac{1}{2} + \frac{m-t}{m} \varepsilon$ otherwise.


Essentially, within a block, arm $1$ starts identical to arm $2$ (with $\bern{\nicefrac{1}{2}}$ rewards), and for the first half of the block, drifts (by a suitably chosen value $\varepsilon$) upwards and away from a mean of $\nicefrac{1}{2}$, and reaches a maximum reward mean gap at half-way point.
Then, for the remainder of the block, it drifts towards (downwards) a mean of $\nicefrac{1}{2}$ and reaches back the $\bern{\nicefrac{1}{2}}$ reward distribution at the end of the block.

With this setup, we argue that when presented with a randomly chosen bandit problem instance
between $\nu$ and $\nu'$, any algorithm is condemned to incur in expectation the stated minimum regret.

\section{NUMERICAL RESULTS}
\label{sec:experiments}
 
\begin{figure*}[!h]
    \centering
    \begin{subfigure}[b]{0.46\textwidth}
            \centering
            \includegraphics[width=\linewidth]{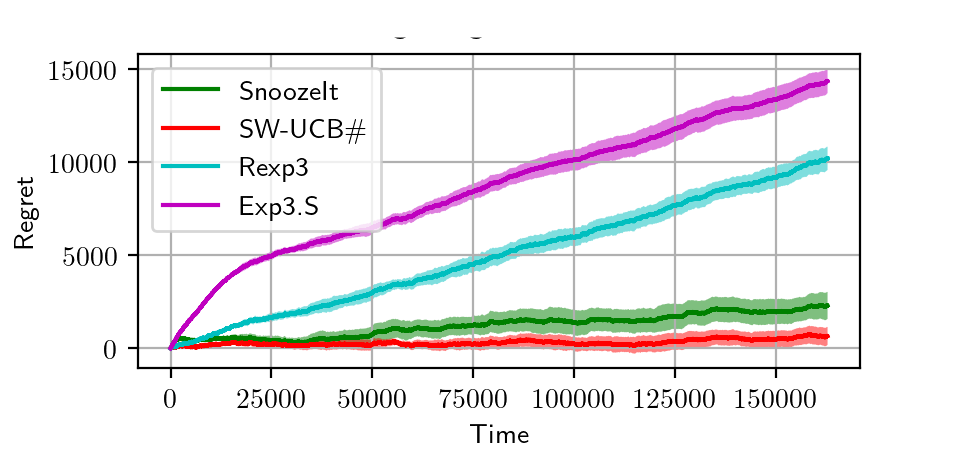}
            \caption{}
            \label{fig:regret-low-delt-well-sep}
    \end{subfigure}%
    \begin{subfigure}[b]{0.4\textwidth}
            \centering
			\includegraphics[width=\linewidth]{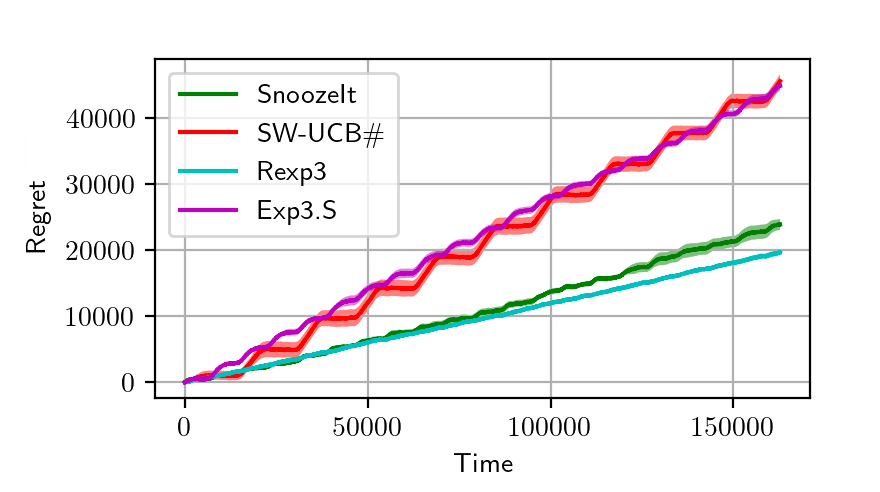}
            \caption{}
            \label{fig:regret-high-delt-oscillating-arms}
    \end{subfigure}
    \caption{Comparison of average regret among the 4 algorithms. The translucent regions (of the same colour) around the curves mark 1 standard deviation over 10 runs. 
    }
\end{figure*}

\begin{figure*}[!h]
    \centering
    \begin{subfigure}[b]{0.47\textwidth}
            \centering
		    \includegraphics[width=\linewidth]{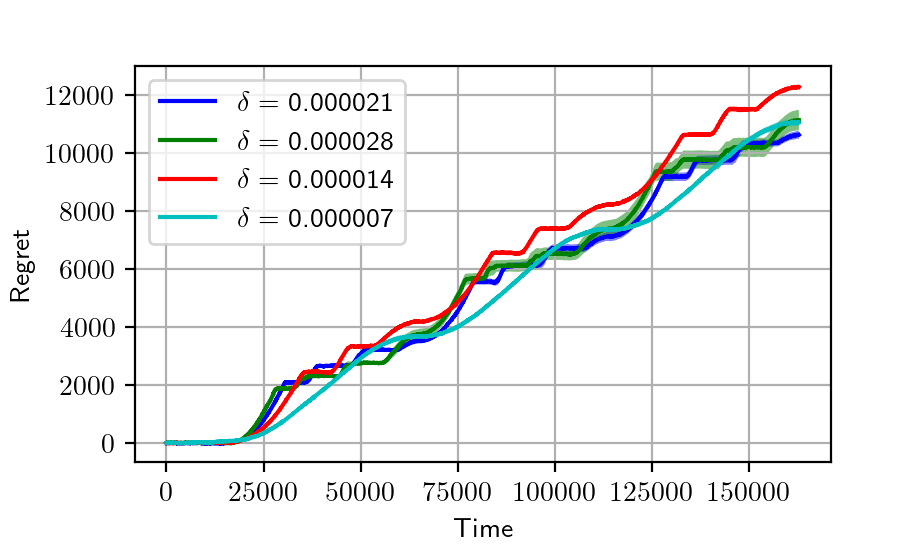}
            \caption{}
            \label{fig:regret-snoozeit-multi-delta-1}
    \end{subfigure}%
    \begin{subfigure}[b]{0.44\textwidth}
            \centering
			\includegraphics[width=\linewidth]{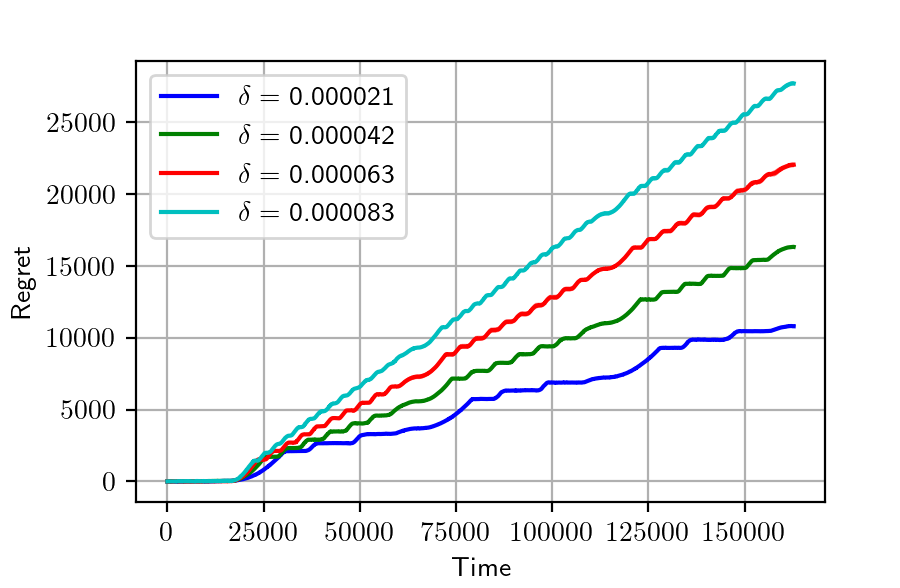}
            \caption{}
            \label{fig:regret-snoozeit-multi-delta-2}
    \end{subfigure}
    \caption{Comparison of average regret incured by $\AlgGE$. The translucent regions (of the same colour) around the curves mark 1 standard deviation over 10 runs. }
\end{figure*}

We numerically evaluate the performance of $\AlgGE$\footnote{
	As noted in Remark \ref{rem:empirical-tweaks}, we slightly modify our algorithm 
	(as in appendix \ref{appn-subsec:modified-algo}) to possibly realise its full potential.} on synthetic problem instances. We also implement algorithms from the literature and conduct a comparative study with different problem instances.
In particular, we consider the following algorithms; 
(1) $\rexp$ \citep{besbes2014stochastic}, that divides the time into blocks upfront and repeatedly runs the Exp3 algorithm 
from scratch in each block,
(2) $\swucb$ \citep{wei2018abruptly}, a soft exploration algorithm based on the arms' reward means' upper confidence bounds that are computed over a sliding window that enlarges with time.
(3) $\exps$ \citep{besbes2019optimal}, a `smoother' variant of Exp3 algorithm.
We note that $\rexp$ and $\exps$ are originally proposed for the more general total variation-budgeted setting.
Also, all these algorithms have knowledge of the drift limit $\delta$ (or the appropriate drift parameter) in advance.

We consider a time horizon of $T\simeq 160,000$, 
and report results averaged from 10 runs. 
The rewards are drawn from a Gaussian distribution with the specified mean, and a variance of \nicefrac{1}{4}.
\footnote{Indeed, the choice of a Gaussian draw with \nicefrac{1}{4} variance fits in with our theoretical analysis, specifically, the Hoeffding's inequality usage in Claim \ref{clm:good-event-prob}}.

First, we run all the algorithms on a problem instance 
with a low value of local change   $\delta=\nicefrac{1}{(10 c_9 \log T)} \simeq 0.000021$  where the arms are well-separated (refer to  fig. \ref{fig:instance-low-delt-well-sep} in appendix \ref{appn-subsec:missing-figures}). The identity of the optimal arm remains unchanged throughout.
We observe (fig. \ref{fig:regret-low-delt-well-sep}) that $\rexp$ and $\exps$ perform poorer compared to
$\swucb$ and $\AlgGE$. 
While $\AlgGE$ and $\swucb$ are very adaptive in their behaviour to the observed empirical reward means,
we see that $\rexp$ predetermines the sizes of blocks (tailored to the drift parameter),
and $\exps$, after tuning the weights, additionally boosts up the weights of the arms equally without taking into account the observed empirical gaps.

Second, we run all 4 algorithms on a problem instance 
 with a relatively larger value of local change $\delta=\nicefrac{1}{( c_9  \log T)} \simeq 0.00021$, where both arms have short alternating stretches of stationarity and drifts, with the identity of the optimal arm toggling with every drift (fig. \ref{fig:instance-high-delt-oscillating-arms} in appendix \ref{appn-subsec:missing-figures}).
Interestingly, we observe (fig. 
\ref{fig:regret-high-delt-oscillating-arms}) that $\swucb$ performs poorer compared to $\AlgGE$. 
We hypothesise that $\AlgGE$ makes better use of short frequent stretches of stationarity; 
however, with the arms oscillating as we have here, when the sliding window encompasses the optimal stretches (and the drifts) of both arms,
$\swucb$ does not take into account the recency of the samples within the sliding window. 
$\AlgGE$, on the other hand, operates with dynamically-sized windows based on the observed detectable gaps that always capture the most suitably sized number of recent samples.

Third, we evaluate the performance of our algorithm on different instances.
In particular, we consider 4 instances with similar structure that differ only by the drift limit $\delta$ imposed, and have common periods of stationarity and drift
(refer fig. \ref{fig:instance-snoozeit-multi-delta-1} in appendix \ref{appn-subsec:missing-figures}).
We observe (fig. \ref{fig:regret-snoozeit-multi-delta-1}) that the average regret does not correlate with an increase in drift limit $\delta$. 
The well separated nature of the arms in instances with a higher $\delta$ improves performance (specifically, increased detected (empirical) gaps lengthen the snooze periods),
and potentially offsets the higher regret causable due to more frequent episodes (specifically, larger $\delta$ values shorten the snooze periods).

Finally, we consider another set of 4 structurally similar instances with different $\delta$, where arms have equal total cumulative drift and almost common periods of stationarity and drift (refer fig. \ref{fig:instance-snoozeit-multi-delta-2} in appendix \ref{appn-subsec:missing-figures}). 
With the same maximum gaps among instances, the regret increases with drift limit $\delta$ (fig. \ref{fig:regret-snoozeit-multi-delta-2}), which is in line with what our theory lays down.

\section{CONCLUSION}


Our characterization of the detectable gap profile enabled an instance-dependent characterization of the regret in the drifting setting. We believe that the detectable gap profile 
is a fundamental property of non-stationary bandit settings that may hold the key to more refined performance analysis, beyond merely the slowly-varying setting considered here. On that note, one interesting direction to pursue would be to characterize instant-dependent regret bounds
in the more general total variation-budgeted setting.

\bibliographystyle{named}
\bibliography{references}

\appendix

\section{Proof of Theorem \ref{thm:inst-dep-reg}}
\label{appn-sec:instance-upper-bound}

We proceed to derive an instance-dependent regret upper bound expression for $\AlgGE$. 
Towards that, we first establish useful properties about
the nature of the active and passive phases, the sizes of episodes,
the connection between the regret and the detectable gap, $\lambda$.

Let $\GoodEvent$ denote the \emph{good} event where for all time-steps and all arms, the empirical reward means for all valid window sizes fall close to the true reward means. In other words,
\begin{align*}
    &\GoodEvent := \left\{ \forall a \in \Arms, t \in [T], \forall w: \muhat_{a,t}(w) - r(w) < \mu_{a,t}(w) < \muhat_{a,t}(w) + r(w) \right\}. \numberthis \label{eqn:good-event}
\end{align*}

\begin{restatable}{claim}{ClaimGoodEvent}
\label{clm:good-event-prob}
    The good event $\GoodEvent$ occurs with high probability; specifically,
    $\prob{\GoodEvent} \geq 1 - \frac{2}{T}$.
\end{restatable}

To retain flow of thought, we defer the proof of the claim to appendix \ref{appn-subsec:good-event-prob}.

For the rest of the analysis in this section until Theorem \ref{thm:inst-dep-reg} is restated, we shall assume that the event $\GoodEvent$ occurs.

\begin{lemma}[Sub-optimality of snoozed arms]
\label{lem:inactive-not-optimal}
    A snoozed arm is never optimal during the passive phase.
    Precisely,
    in any episode $i \in [e]$,
    for active arms, $a$ and $b$, 
    such that $a \lambTilGreat b$ at time $g_i$, 
    for all times $t \in [g_i+1, t_{i+1}]$, we have $\mu_{a,t} \geq \mu_{b,t}$.
\end{lemma}
\begin{proof}
    Note that if there is no passive phase in the episode, i.e., arm $b$ is not snoozed, 
    then the result is vacuously true. Thus, what is to be shown is only the case wherein arm $b$ is snoozed, 
    equivalently, there is a non-empty passive phase $[g_i +1, t_{i+1}]$.
    We prove this lemma in two steps. 
    First, we show (in Claim \ref{clm:recent-gap-existance}) that when $\AlgGE$ detects 
    $a \lambTilGreat b$ at time $g_i$,
    then, for a period of time culminating at $g_i$, 
    arm $a$ has had a larger average true reward mean than that of arm $b$
    by a certain margin.
    Second, we argue that the duration of the passive phase for which $b$ is snoozed 
    is chosen based on this margin such that arm $b$ remains sub-optimal
    for the entirety of the passive phase.

    At the end of the active phase of episode $i$, at $g_i$, 
    $\AlgGE$ detects that for some $\lambTil \in [0,1]$,
    $a \lambTilGreat b$.
    We make the following claim about the gap between the true reward means of the two arms 
    at some point in time during this active phase.
    
    \begin{restatable}{claim}{ClaimRecentGapExistance}
    \label{clm:recent-gap-existance}
        There exists a time-step $t' \in [t_i + 1, g_i]$
        such that $\mu_{a,t'} - \mu_{b,t'} \geq \frac{\lambTil}{3} - 2\delta \geq 4 \sqrt{\frac{\log T}{\tau_i}} - 2\delta$.
    \end{restatable}
    \begin{proof}
        As arm $a$ is $\lambTil$-better than arm $b$ at time $t = g_i$, 
        by Definition \ref{def:lambda-better}, we have, for a window sample size of $w := \ceil{\dfrac{c_1 \log T}{\lambTil^2}}$, with $c_1 = 72$,
        \begin{align*}
            & LCB_{a,t}(w) > UCB_{b,t}(w) + 2r(w) -\delta \\
            \implies & \muhat_{a,t}(w) - r(w) > \muhat_{b,t}(w) + r(w) + 2r(w) -\delta\\
            \implies & \mu_{a,t}(w) > \muhat_{a,t}(w) - r(w) > \muhat_{b,t}(w) + r(w) + 2r(w) - \delta > \mu_{b,t}(w) + 2r(w) -\delta \tag{$\because$ event $\GoodEvent$ occurs}\\
            \implies & \mu_{a,t}(w) - \mu_{b,t}(w) > 2 r(w) -\delta\\
            \implies & \mu_{a,t}(w) - \mu_{b,t}(w) > 2 \sqrt{\frac{2 \log T}{w}} -\delta = 2 \sqrt{\frac{2 \log T . \lambTil^2}{72 \log T}} -\delta = 2 \sqrt{\frac{ \lambTil^2}{36}} -\delta = \frac{\lambTil}{3} -\delta.
            \numberthis \label{eqn:recent-gap-lambda}
        \end{align*}
        
        Note that $\lambTil \in [0,1]$ is constrained (again, as in Definition \ref{def:lambda-better}) by 
        the number of samples $w$ available in the current episode $i$ as follows.

        \begin{align*}
            &\ceil{\frac{c_1 \log T}{\lambTil^2}} \leq \frac{\tau_i}{2} \\
            \implies & \frac{2 c_1 \log T}{\tau_i} \leq \lambTil^2 \\
            \implies & \lambTil \geq \sqrt{\frac{2 c_1 \log T}{\tau_i}}. \numberthis \label{eqn:lambtil-gauge-curve-comparison}
        \end{align*}

        We use the short-hand $f := \frac{1}{3}\sqrt{\frac{2 c_1 \log T}{\tau_i}} = 4 \sqrt{\frac{\log T}{\tau_i}}$ for ease of expression.

        In both terms $\mu_{a,t}(w)$ and $\mu_{b,t}(w)$, 
        each of the $w$ time-steps whose reward means are averaged
        is from the time period $[t - 2w + 1,t]$. 
        Denote $s=t-2w+1$ (note $s > t_i$) to be the  time-step of the earliest of the samples considered in the statistical test. 
        Expanding the reward mean ($\mu$) terms and
        using $f \leq \frac{\lambTil}{3}$  (from \ref{eqn:lambtil-gauge-curve-comparison})
        in Equation \ref{eqn:recent-gap-lambda}, we get

        \begin{equation*}
            \mu_{a,t}(w) - \mu_{b,t}(w) = 
            \frac{1}{w} \sum_{i=0}^{w-1} \mu_{a,s + 2i} - \mu_{b,s + 2i + 1}
            \footnote{Assume w.l.o.g. that sampling starts with arm $a$, alternates between the two arms, and ends with arm $b$.}
            > \frac{\lambTil}{3} -\delta \geq f -\delta. \numberthis \label{eqn:lambtil-gauge-slack}
        \end{equation*}
        
        The average of $w$ terms being larger than $f$
        implies that atleast one term is larger than $f$.
        Thus, for some $0 \leq i \leq w-1$, we have $\mu_{a,s + 2i} - \mu_{b,s + 2i + 1} > f - \delta$.
        Also with the drift limit implication as in Equation \ref{eqn:drift-limit},
        we have $\mu_{a,s + 2i} - \mu_{b,s + 2i} > f - 2 \delta$.
        Thus, we get for some $t' \in [s,t] \subseteq [t_i+1,g_i]$ 
        that $\mu_{a,t'} - \mu_{b,t'} \geq f - 2 \delta$.
        The claim follows.
    \end{proof}


    We continue to use the short-hand $f := 4 \sqrt{\frac{\log T}{\tau_i}}$ for ease of expression.
    From Claim \ref{clm:recent-gap-existance}, 
    at some time $t' \in [t_i+1 , g_i]$, we have 
    $\mu_{a,t'} - \mu_{b,t'} > f - 2 \delta$.
    By adhering to the drift limit, we have $\abs{\Delta_t - \Delta_{t+1}} \leq 2 \delta$ for all time-steps $t \in [T-1]$.
    Thus, for the identity of the optimal arm to change, i.e., for the gap of $f - 2 \delta$ to get exhausted, it requires $\nicefrac{(f-2\delta)}{2\delta} = \nicefrac{f}{2\delta} - 1$ time-steps.
    Thus, for all $t'' \in [t', t' + \nicefrac{f}{2\delta} - 1]$, 
    and thus for all $t'' \in [g_i+1, t_i + \nicefrac{f}{2\delta}]$
    (which is a shorter period)
    we have $\mu_{a,t''} > \mu_{b,t''}$.
    i.e., arm $b$ is sub-optimal in the time period 
    $[g_i+1, t_i + \nicefrac{f}{2 \delta}]$.

    Once $\AlgGE$ detects $a \lambTilGreat b$ at $t$, 
    it computes the duration of the sub-optimality buffer period as 
    $\text{buf} = \nicefrac{f}{2 \delta}$ (in Line \ref{line:compute-buffer}).
    If $\text{buf} \leq \tau_i$, no arm is snoozed for a passive phase.
    Otherwise, as in Line \ref{line:snooze}, 
    the sub-optimal arm $b$ is snoozed until time-step
    $ t_{i+1} = t_i + \text{buf} = t_i + \nicefrac{f}{2 \delta}$, 
    i.e., the passive phase runs for the period $[g_i+1, t_i + \nicefrac{f}{2 \delta}]$. 
    We have already shown that during this period arm $b$ is sub-optimal.
\end{proof}

Thus, thanks to Lemma \ref{lem:inactive-not-optimal}, during the passive phase of an episode, only the optimal arm is played.

Based on the above analysis so far, we slightly step aside from the proof to make the following remark about how the algorithm can be tightened without compromising on the theoretical guarantees.





\begin{remark}
\label{rem:empirical-tweaks}
    In the design of $\AlgGE$, we make the following observations
    \begin{enumerate}
        \item The snooze end time $t_i + \text{buf}$ is conservative, and can be pushed further (into the future) to $g_i - 2w + \text{buf}$. This is because Claim \ref{clm:recent-gap-existance} holds even for a tighter time period of $[g_i-2w+1, g_i]$ and it doesn't affect the correctness of Lemma \ref{lem:inactive-not-optimal}.
        \item The computed buffer time $\text{buf}=\nicefrac{f}{2 \delta}$ can be increased to $\nicefrac{\lambTil}{6 \delta}$ without affecting the correctness of the Lemma. It merely cuts the slack in equation \ref{eqn:lambtil-gauge-slack}.
    \end{enumerate} 
    Both these modifications shall possibly lead to a smaller regret. However, for our theoretical analysis, the current version is sufficient and in fact, it eases the anlaysis considerably.
    Nevertheless, we do incorporate these modifications in usage for the experiments (in Section \ref{sec:experiments}).
\end{remark}

We now bound the expected regret incurred by the sub-optimal arm, which can be different arms in different stretches of time, during the active phase of episode $i$. 
Towards that, we define the notion of regret in an episode:

\begin{definition}[Episode Regret]
    In an algorithmic run of $\AlgGE$ with a drift limit parameter $\delta$ on a problem instance $\mu \in S_\delta$, 
    the regret in any episode $i \in [e]$ is defined as 
    \begin{align*}
        R_i = \sum_{t=t_i + 1}^{t_{i+1}} \mu^*_t - \mu_{\AlgGE(t), t}.
    \end{align*}
    where $\mu^*_t = \max_{a \in \Arms} \mu_{a,t}$ is the mean reward of the optimal arm at time $t$,
    and $\AlgGE(t) \in \Arms$ is the arm pulled by $\AlgGE$ at time $t \in [T]$.
\end{definition}


Next, we upper bound the regret $R_i$ in episode $i$ based on $g_i$, 
the time at which the statistical test passes (or the active phase of the episode ends).  
Towards that, we state and prove a key lemma.
\begin{lemma}
\label{lem:episodic-regret-bound}
    In an algorithmic run of $\AlgGE$ with a drift limit parameter $\delta$ on a problem instance $\mu \in S_\delta$, 
    the regret in episode $i \in [e]$ is upper bounded as
    \begin{align*}
        R_i \leq c_4. \dfrac{1}{\lambda_{g_i - 1}} \log T + c_5,
    \end{align*}
    where $c_4 = 72, c_5=1$ are constants.
\end{lemma}
\begin{proof}
    We split the regret in episode $i$ into two components;
    the regret incurred in the active phase of episode $i$, 
    and the regret incurred in the passive phase of episode $i$:


    \begin{align*}
        R_i = \sum_{t=t_i + 1}^{g_i} \mu^*_t - \mu_{\AlgGE(t), t} + \sum_{g_i + 1}^{t_{i+1}} \mu^*_t - \mu_{\AlgGE(t), t}.
    \end{align*}

    If there is no passive phase in episode $i$, then $g_i = t_{i+1}$, and the second term is trivially $0$.
    Otherwise, 
    by Lemma \ref{lem:inactive-not-optimal} 
    we have that the snoozed arm remains sub-optimal for the entire passive phase, i.e., at every time-step in the passive phase, the optimal arm at that time is played.
    Thus, the second term equals $0$.

    Also, we trivially upper bound the regret at time $g_i$ by $1$ to get
    \begin{align*}
        R_i = 1 + \sum_{t=t_i + 1}^{g_i - 1} \mu^*_t - \mu_{\AlgGE(t), t}. \numberthis \label{eqn:episodic-regret-1}
    \end{align*}

    Recall that $\AlgGE$ alternates between the two arms in the time period $[t_i + 1, g_i - 1]$, 
    and the statistical test did not pass at time $g_i - 1$, i.e.,
    for the two arms, $a$ and $b$, there was no $\lambTil \in [0,1]$ 
    (and a corresponding $w:=\ceil{\frac{c_1 \log T}{\lambTil^2}}$) 
    for which $a \lambTilGreat b$ at $g_i - 1$.

    Applying Definition \ref{def:lambda-better} at time $t=g_i - 1$ with a choice of $w=\frac{\tau_i-1}{2}$ 
    \footnote{For technical clarity, we assume integrality of all quantities suitably.} (for a suitable $\lambTil$), we have

    \begin{align*}
        LCB_{a,t}(w) \leq& UCB_{b,t}(w) + 2 r(w) - \delta\\
        \implies \muhat_{a,t}(w) - r(w) \leq& \muhat_{b,t}(w) + r(w) + 2 r(w) - \delta\\
        \implies \mu_{a,t}(w) - r(w) -r(w) \leq \muhat_{a,t}(w) - r(w) \leq& \muhat_{b,t}(w) + r(w) + 2 r(w) - \delta \leq \mu_{b,t}(w) + r(w) + r(w) + 2 r(w) - \delta\\
        \implies \mu_{a,t}(w) - \mu_{b,t}(w) \leq& 6 r(w) - \delta\\
        \iff \mu_{a,t}\left(\tfrac{\tau_i-1}{2}\right) - \mu_{a,t}\left(\tfrac{\tau_i-1}{2}\right) 
        \leq& 6 \sqrt{\dfrac{2\log T}{\tfrac{\tau_i-1}{2}}} - \delta\\
        \iff \mu_{a,t}\left(\tfrac{\tau_i-1}{2}\right) - \mu_{b,t}\left(\tfrac{\tau_i-1}{2}\right)
        \leq& 12 \sqrt{\dfrac{\log T}{\tau_i - 1}} - \delta. \numberthis \label{eqn:max-ave-gap}
    \end{align*}

    The third is due to occurrence of the good event $\GoodEvent$.
    By expanding the $\mu_{a,t}, \mu_{b,t}$ terms further 
    with window size $w = \frac{\tau_i - 1}{2}$, 
    and a corresponding starting time of samples, $s=t_i+1$,
    we have\footnote{Assume w.l.o.g. that sampling starts with arm $a$, alternates between the two arms, and ends with arm $b$.}

    \begin{align*}
        &\mu_{a,t}(w) - \mu_{b,t}(w)  \\
        = &  \frac{1}{w} \sum_{i=0}^{w-1} \mu_{a,s + 2i} - \mu_{b,s + 2i + 1} \\
        \geq & \frac{1}{w} \sum_{i=0}^{w-1} \mu_{a,s + 2i + 1} - \mu_{b,s + 2i + 1} - \delta \\
        = & \frac{1}{w} \left( -w \delta + \sum_{t=t_i+1}^{g_i -1} \mu^*_t - \mu_{\AlgGE(t), t} \right).
        \numberthis \label{eqn:max-ave-gap-to-reg}
    \end{align*}

    The final equality is due to the alternating nature of the arm plays.
    Combining inequalities \ref{eqn:max-ave-gap} and \ref{eqn:max-ave-gap-to-reg} yields

    \begin{align*}
        \frac{1}{w} \left( -w \delta + \sum_{t=t_i+1}^{g_i -1} \mu^*_t - \mu_{\AlgGE(t), t} \right) &\leq 12 \sqrt{\dfrac{\log T}{\tau_i - 1}} - \delta \\
        \implies -w \delta + \sum_{t=t_i+1}^{g_i -1} \mu^*_t - \mu_{\AlgGE(t), t} &\leq w.12 \sqrt{\dfrac{\log T}{\tau_i - 1}} - w \delta \\
        \implies \sum_{t=t_i+1}^{g_i -1} \mu^*_t - \mu_{\AlgGE(t), t} &\leq \frac{\tau_i-1}{2}.12 \sqrt{\dfrac{\log T}{\tau_i - 1}} \\
        & = 6 \sqrt{(\tau_i - 1) \log T}.
    \end{align*}

    Substituting this in eqn. \ref{eqn:episodic-regret-1}, we have
    \begin{align*}
        R_i &\leq  1 + 6 \sqrt{(\tau_i-1) \log T}. \numberthis \label{eqn:episodic-regret-2}
    \end{align*}

    Next, we upper bound the term $\sqrt{\tau_i-1}$ in terms of the detectable gap profile $\lambda$, an instance-dependent quantity.

    \begin{lemma}
    \label{lem:gauge-lambda-curve-relation}
        For some $\tau > 0$, if in the active phase of episode $i$, at time $t' = t_i + \tau$, we have $\lambda_{t'} > \sqrt{\frac{c_2 \log T}{\tau}}$,
        then the statistical test passes, i.e., at time $t'$, for two arms $a,b$, we have $a \lambTilGreat b$.
        Here, $c_2 = 144$ is a constant.
    \end{lemma}
    \begin{proof}
        If $\lambda_{t'} = \sqrt{\frac{c_2 \log T}{t'}}$ (from second assignment in eqn. \ref{eqn:lamdba-curve}), we always have
        $\lambda_{t'} \leq \sqrt{\frac{c_2 \log T}{\tau}}$ as $\tau \leq t'$. 
        But, given that we have the opposite assumption in the Lemma statement, we can make the following claim from the description of detectable gap $\lambda$ (first assignment in eqn. \ref{eqn:lamdba-curve}).

        \begin{claim}
        \label{clm:lambda-implication-on-mus}
            The average of the true reward means of arms $a$ and $b$ at time $t'$ in a window of sample size $w:=\frac{c_1 \log T}{\lambda_{t'}^2}$ (or $w = \frac{w(\lambda_{t'})}{2}$) obeys
            $\mu_{a,t'}(w) -  \mu_{b,t'}(w) \geq \lambda_{t'} - \delta$. 
            Here, $c_1=72$ is a constant.
        \end{claim}
        \begin{proof}
            From defn. \ref{eqn:lamdba-curve}, we have for some $\alpha$, for starting time $s=t'-2w+1$,
            \begin{align*}
                \frac{1}{2w} \sum_{t=s}^{t'} \mu_{a,t} - \mu_{b,t} = \alpha \geq \lambda_{t'}. \numberthis \label{eqn:lamb-implic-1}
            \end{align*}

            Also, by the drift limit implication, we have
            \begin{align*}
                \frac{1}{w} \sum_{i=0}^{w-1} \mu_{a,s+2i+1} - \frac{1}{w} \sum_{i=0}^{w-1} \mu_{a,s+2i} &\leq \delta, \quad \text{and} \numberthis \label{eqn:lamb-implic-2}\\
                \frac{1}{w} \sum_{i=0}^{w-1} \mu_{b,s+2i+1} - \frac{1}{w} \sum_{i=0}^{w-1} \mu_{b,s+2i} &\leq \delta. \numberthis \label{eqn:lamb-implic-3}
            \end{align*}
            By subtracting eqn. \ref{eqn:lamb-implic-2} from eqn. \ref{eqn:lamb-implic-1}, and subtracting eqn. \ref{eqn:lamb-implic-3} from eqn. \ref{eqn:lamb-implic-1}, and adding up the two resultant inequalities, we get
            \begin{align*}
                \frac{1}{w} \left(\sum_{t=s-2w+1}^{t'} \mu_{a,t} - \mu_{b,t} \right) + \frac{2}{w} \sum_{i=0}^{w-1} \mu_{a,s+2i} - \frac{2}{w} \sum_{i=0}^{w-1} \mu_{b,s+2i+1} &\geq 4 \alpha - 2 \delta \\
                \implies \frac{1}{w} \sum_{i=0}^{w-1} \mu_{a,s+2i} - \frac{1}{w} \sum_{i=0}^{w-1} \mu_{b,s+2i+1} &\geq \lambda_{t'} - \delta \\
                \iff \mu_{a,t'}(w) -  \mu_{b,t'}(w) &\geq \lambda_{t'} - \delta. 
            \end{align*}
            This proves the claim.
        \end{proof}

        By the hypothesis in the lemma, when $\lambda_{t'} > \sqrt{\frac{c_2 \log T}{\tau}}$, or equivalently $\tau > \frac{144 \log T}{\lambda_{t'}^2}$,
        we have that $\tau > 2w$, thus, $w$ is a valid window sample count at time $t$
        for which the assumption of \emph{good event}, $\GoodEvent$,
        that bounds the deviation of empirical means from the true means, can be used.
        By applying eqn. \ref{eqn:good-event} to the conclusion of Claim \ref{clm:lambda-implication-on-mus}, we have
        \begin{align*}
            \muhat_{a,t'}(w) - \muhat_{b,t'}(w) &\geq \lambda_{t'} - \delta - 2 r(w) \\
            \implies LCB_{a,t'}(w) - UCB_{a,t'}(w) &\geq \lambda_{t'} - 4 r(w) -\delta \\
             &\geq \sqrt{\frac{72 \log T}{w}} - 4 \sqrt{\frac{2 \log T}{w}} -\delta  \\
             &= 2 \sqrt{\frac{2 \log T}{w}} = 2 r(w) -\delta \numberthis \label{eqn:lambda-better-sufficiency} \\
        \end{align*}
        The inequality in equation \ref{eqn:lambda-better-sufficiency} is the condition for
        the algorithm to declare $a \lambTilGreat b$ 
        (as in Definition \ref{def:lambda-better}) at $t'$.
        Thus, the statistical test passes.
    \end{proof}

    Now, recall that the statistical test did not pass at $g_i-1 = t_i+\tau_i-1$. 
    By Lemma \ref{lem:gauge-lambda-curve-relation}, we have that $\lambda_{g_i - 1} \leq \sqrt{\frac{c_2 \log T}{\tau_i - 1}}$,
    or equivalently, $\sqrt{\tau_i-1} \leq \frac{\sqrt{c_2 \log T}}{\lambda_{g_i - 1}}$.
    Substituting this in inequality \ref{eqn:episodic-regret-2}, the lemma follows:

    \begin{align*}
        R_i &\leq 1 + \leq 6 \frac{\sqrt{c_2 \log T}}{\lambda_{g_i - 1}} \sqrt{\log T} \\
        &= 72 \dfrac{1}{\lambda_{g_i - 1}} \log T + 1 \\
        &= c_4 \dfrac{1}{\lambda_{g_i - 1}} \log T + c_5.
    \end{align*}
\end{proof}

We show an instance-dependent regret bound over the entire time horizon 
as an accumulation of episode-wise regret bounds.
To show this upper bound, 
we partition the time-horizon into \emph{blocks} of suitably small size, 
and bound the regret in episodes by accounting each of them to some block.
Finally, we shall condition our regret upper bounds on occurrence of events $\GoodEvent$ and $\GoodEvent'$ 
to arrive at the final instance-dependent upper bound.

Recall an algorithmic run of $\AlgGE$ has $e$ episodes, indexed $1,2,\dots,e$,
which start after times $t_1 = 0, t_2, \dots, t_e$ (and a hypothetical $t_{e+1}=T$ marks the end of the last episode $e$) respectively.
Note that the collection of episode time periods $\left([t_i + 1, t_{i+1}]\right)_{i \in [e]}$ partitions the entire time horizon $[T]$.
Thus, the total regret $R(\AlgGE) = \sum_{i \in [e]} R_i$. 

For analysing the total regret, we partition the time horizon into $m = \nicefrac{T}{\tau}$ blocks ($1,2, \dots, m$), each 
of size $\tau = c_3 \delta^{-\nicefrac{2}{3}} \log^{\nicefrac{1}{3}} T$, where $c_3 = 2^{2/3}$ is a constant. 
Each block $j \in [m]$ spans the time period $b_j := [(j-1)\tau + 1, j\tau] \cap [T]$.
This choice of the size of a block is based on the minimum duration of an episode we shall establish below.

\begin{lemma}
\label{lem:size-of-episode}
    In an algorithmic run of $\AlgGE$, the duration of any episode $i \in [e-1]$ (except the last one), 
    $t_{i+1} - t_i \geq c_3 \delta^{\nicefrac{-2}{3}} \log ^{\nicefrac{1}{3}} T$.
    Here, $c_3 = 2^{\nicefrac{2}{3}}$ is a constant.
\end{lemma}
\begin{proof}
    At the end of the active phase (of duration $\tau_i$) at time-step $g_i$, the sub-optimality buffer computed is $\text{buf} = \frac{f}{2\delta} = \frac{2}{\delta} \sqrt{\frac{\log T}{\tau_i}}$.
    In Lines \ref{line:check-snooze-condition}-\ref{line:snooze}, the algorithm decides what the snooze duration of the sub-optimal arm should be, 
    thus determining $t_{i+1}$, the time when the episode $i$ shall end, based on the following two cases.
    
    \paragraph{Case $\tau_i \geq \text{buf}$ :} Equivalently, we have $\tau_i \geq \frac{2}{\delta} \sqrt{\frac{\log T}{\tau_i}}$.
    The sub-optimal arm does not get snoozed, so, episode $i$ terminates at $t_{i+1} = g_i = t_i + \tau_i$.
    
    \paragraph{Case $\tau_i < \text{buf}$ :} Equivalently, we have $\tau_i < \frac{2}{\delta} \sqrt{\frac{\log T}{\tau_i}}$.
    The sub-optimal arm gets snoozed until time $t_i + \text{buf} 
     = t_i + \frac{2}{\delta} \sqrt{\frac{\log T}{\tau_i}}$. 
    The end of the passive phase marks the end of episode $i$,
    thus, $t_{i+1} = t_i + \frac{2}{\delta} \sqrt{\frac{\log T}{\tau_i}}$

    From the two cases, we can see that $t_{i+1} = t_i + \max \left\{\tau_i, \frac{2}{\delta} \sqrt{\frac{\log T}{\tau_i}} \right\}$.
    To lower bound $t_{i+1} - t_i$, the length of episode $i$, 
    we minimize $\max\left\{\tau_i, \frac{2}{\delta} \sqrt{\frac{\log T}{\tau_i}} \right\}$. 
    As the two quantities grow oppositely with $\tau_i$, the minimum occurs when
    \begin{align*}
        \tau_i = \dfrac{2}{\delta} \sqrt{\frac{\log T}{\tau_i}} \quad \Leftrightarrow \quad \tau_i = 2^{\nicefrac{2}{3}} \delta^{\nicefrac{-2}{3}} \log^{\nicefrac{1}{3}} T.
    \end{align*}

    Thus, we have $t_{i+1} - t_i \geq c_3  \delta^{\nicefrac{-2}{3}} \log^{\nicefrac{1}{3}} T$, a lower bound on the duration of an episode.
\end{proof}

From Lemma \ref{lem:size-of-episode}, within a time horizon of size $T$, for later use,
we bound the maximum number of episodes in the following corollary.

\begin{corollary}
\label{cor:max-number-episodes}
    In an algorithmic run of $\AlgGE$, the number of episodes, $e$, is bounded as 
    $e \leq c_3^{-1} T \delta^{\nicefrac{2}{3}} \log ^{\nicefrac{-1}{3}} T$, where $c_3 = 2^{\nicefrac{2}{3}}$ is a constant.
\end{corollary}

As stated earlier, the chosen size of a block is the minimum duration of an episode stated in Lemma \ref{lem:size-of-episode}.
Thus, a block period overlaps with a maximum of two episodes.
Thus, for a maximum of two episodes, say $i,i+1$, 
the time-steps before the statistical tests pass ($g_i - 1$ and $g_{i+1} - 1$)
fall within a block $j$, i.e., $g_i - 1, g_{i+1} - 1 \in b_j$.
We account the corresponding episodic regrets $R_i,R_{i+1}$ to block $j$.
We denote the regret to block $j \in [m]$ as $R(j) \leq R_i + R_{i+1}$ 
(indeed, we would have different episodes $i$ for different blocks $j$.)
Note that, every episodic regret is accounted to some block, and every block accounts for at most $2$ episodic regrets. 
We liberally upper bound the regret expression as follows:
\begin{align*}
    R(\AlgGE) \leq \sum_{j \in [m]} R(j). \numberthis \label{eqn:instance-regret-blocks-1}
\end{align*}

\begin{claim}
\label{clm:block-regret-bound}
    The regret accounted to a block $j \in [m]$ is 
    \begin{align*}
        R(j) \leq c_6. \frac{1}{\lambda_{min}(j)} + c_7
    \end{align*}
    where $\lambda_{min}(j) = \min_{t \in b_j}\lambda_t$.
    Here, $c_6 = 144, c_7=2$ are constants.
\end{claim}
\begin{proof}
    Towards proving this, we upper bound the regret of an episode $i$, $R_i$, that is accounted to block $j$
    and multiply that by two as possibly the regret in another episode ($R_{i+1}$) is also accounted to block $j$.

    By the accounting criteria, we have that $g_i - 1 \in b_j$.
    From Lemma \ref{lem:episodic-regret-bound}, we have $R_i \leq c_4. \dfrac{1}{\lambda_{g_i - 1}} \log T + c_5$.
    As $g_i-1$ could be any time in the range $b_j$, we take the maximum regret over all possile times in $b_j$.
    That is,
    \begin{align*}
        R_i \leq& \max_{t \in b_j} c_4 \dfrac{1}{\lambda_t} \log T + c_5.
    \end{align*}

    Thus, we can bound $R(j)$ as follows:
    \begin{align*}
        R(j) \leq& 2 \max_{t \in b_j} c_4 \dfrac{1}{\lambda_t} \log T + 2 c_5 \\
            =& 2 c_4 \dfrac{1}{ \min_{t \in b_j} \lambda_t} \log T + 2 c_5 \\
            =& c_6 \dfrac{1}{\lambda_{min}(j)} \log T + c_7.
    \end{align*}
    This completes the proof.
\end{proof}

Substituting the conclusion of the claim in expression \ref{eqn:instance-regret-blocks-1}, we have
\begin{align*}
    R(\AlgGE) \leq \sum_{j=1}^{m} c_6 \frac{1}{\lambda_{min}(j)} + c_7. \numberthis \label{eqn:instance-regret-blocks-2}
\end{align*}

For the remainder of the proof, we drop the implicit assumption that $\GoodEvent$ occurs,
and upper bound the conditional expected regret of our algorithm $\AlgGE$ when 
high probable event $\GoodEvent$ occurs using expression \ref{eqn:instance-regret-blocks-2},
and then generously upper bound the conditional regret when $\GoodEvent'$ occurs by $T$ (a regret of $1$ for every time-step).
Along with Claim \ref{clm:good-event-prob}, we have
\begin{align*}
    &\expectation{R(\AlgGE)} \\
     =& \expectation{R(\AlgGE | \GoodEvent)}. \prob{\GoodEvent} + \expectation{ R(\AlgGE | \GoodEvent')}. \prob{\GoodEvent'} \\
    \leq & \left( \sum_{j=1}^{m} c_6. \frac{1}{\lambda_{min}(j)} + c_7 \right) + T. \frac{2}{T} \numberthis \label{eqn:bad-event-constant-regret}\\
    =& c_8 + c_6 \sum_{j=1}^{m} \frac{1}{\lambda_{min}(j)}. \log T + c_7,
\end{align*}
where $c_6=144, c_7=2$ (as defined in Claim \ref{clm:block-regret-bound}), and $c_8 = 2$ are constants.

This leads to the Theorem.

\TheoremInstanceDependentUpperBound*

\subsection{Proof of Claim \ref{clm:good-event-prob}}
\label{appn-subsec:good-event-prob}

\ClaimGoodEvent*

\begin{proof}
    
    We work in the probability space in which the rewards from arm pulls $\muhat_{a,t}$s are generated ahead of time, and for every time $t \in [T]$, arm $a \in \Arms$, we have $\muhat_{a,t}$ is an independent sample from $\bern{\mu_{a,t}}$.

    For some arm $a$, for some time-step $t$, for some window  size $w \leq \nicefrac{t}{2}$,
    consider two sequences of Bernoulli random variables with means 
    $\{\mu_{a,t-2i}\}_{i=0}^{w-1}$ and $\{\mu_{a,t-2i-i}\}_{i=0}^{w-1}$.
    The first (resp. second) sequence contains the reward mean from $w$ alternately picked time-steps culminating at $t$ (resp. $t-1$).
    Without loss of generality, consider only the first sequence
    \footnote{The second sequence $\{\mu_{a,t-2i-i}\}_{i=0}^{w-1}$ is actually covered in the corresponding argument for same window size $w$ with current time $t-1$.}, and 
    denote by $\mu_{a,t}(w) = \frac{1}{w} \sum_{i=0}^{w-1} \mu_{a,t-2i}$
    the average of true reward means of this sequence
    .

    Now, the probability of the average of empirical reward means $\muhat_{a,t}(w)$
    deviating from the average of true reward means $\mu_{a,t}(w)$ by more than a radius $r(w) := \sqrt{\frac{2 \log T}{w}}$ is upper bounded using Chernoff-Hoeffding inequality as follows.
    \begin{align*}
        \prob{\abs{\muhat_{a,t}(w) - \mu_{a,t}(w)} \geq r(w)} 
        \leq 2e^{-2w . r(w)^2} 
        = 2e^{-2w. \frac{2 \log T}{w}} 
        = \frac{2}{T^4}.
    \end{align*}
    
    Taking a union bound over all arms $a \in \Arms$ ($\cardinality{\Arms} < T$), all time steps $t \in [T]$, all possible valid play counts $w$ ($w \leq \nicefrac{t}{2} \leq T$) gives
    \begin{align*}
        &\prob{\exists a \in \mathcal{A}, t \in [T], w, \text{ s.t. } 
        \abs{\muhat_{a,t}(w) - \mu_{a,t}(w)} \geq r(w)} \leq \frac{2}{T} \\
        \implies & \prob{\forall a \in \mathcal{A}, t \in [T], \forall w : \abs{\muhat_{a,t}(w) - \mu_{a,t}(w)} \leq r(w)} \geq 1 - \frac{2}{T}.
    \end{align*}
\end{proof}

\section{Proof of Theorem \ref{thm:minimax-bound}}
\label{appn-sec:minimax-upper}

In this section, we show a minimax (instance-independent) upper bound
for the regret incurred by $\AlgGE$
that depends on the time horizon $T$, the drift limit $\delta$,
but, is independent of the actual reward mean profile $\mu$.
Towards this, for every episode, we upper bound the \emph{average regret} or per-time-step regret.
Then, we shall show a minimax regret that is bounded by
$T$ times the maximum average regret incurred in any episode.

\TheoremMinimaxUpperBound*

\begin{proof}
\label{proof:minimiax-upper}
    As shown in equation \ref{eqn:bad-event-constant-regret}, 
    the conditional regret when $\GoodEvent'$ occurs can be bounded as 
    $\expectation{ R(\AlgGE | \GoodEvent')}. \prob{\GoodEvent'} \leq 2$.
    Thus, it is sufficient to prove the required bound under the assumption of occurrence of $\GoodEvent$. 

    From equation \ref{eqn:episodic-regret-2}, the regret in episode $i$ is

    \begin{align*}
        R_i \leq& 1 + 6 \sqrt{(\tau_i -1). \log T} \\
            < & 1 + 6 \sqrt{\tau_i \log T} \numberthis \label{eqn:minimax-exp-1}
    \end{align*}

    The total regret over the entire time horizon is 
    \begin{align*}
        R(\AlgGE) &=  \sum_{i \in [e]} R_i 
                \leq \sum_{i \in [e]} 1 + 6 \sqrt{\tau_i \log T}  &\left\{\text{From eqn. } \ref{eqn:minimax-exp-1} \right\} \\
                &=  \sum_{i \in [e]} 1 + \sum_{i \in [e]} 6 \sqrt{\tau_i \log T} \\
                &\leq  c_3^{-1} T \delta^{\nicefrac{2}{3}} \log ^{\nicefrac{-1}{3}} T + 
            \sum_{i \in [e]} 6 \sqrt{\tau_i \log T} &
            \left\{\text{From corollary } \ref{cor:max-number-episodes} \right\} \\
                &\leq  O(T \delta^{\nicefrac{1}{3}} \log ^{\nicefrac{1}{3}} T) + \sum_{i \in [e]} 6 \sqrt{\tau_i \log T}  \\
                &\leq  O(T \delta^{\nicefrac{1}{3}} \log ^{\nicefrac{1}{3}} T) + T . \max_{i \in [e]} \frac{6 \sqrt{\tau_i \log T}}{t_{i+1} - t_i} \\
                &=  O(T \delta^{\nicefrac{1}{3}} \log ^{\nicefrac{1}{3}} T) + T . \max_{i \in [e]} R^{ave}_i \numberthis \label{eqn:minimax-exp-2} \\
    \end{align*}

    We denote $R^{ave}_i = \frac{R_i-1}{t_{i+1} - t_i} \leq \frac{6 \sqrt{\tau_i \log T}}{t_{i+1} - t_i}$
    to be the average regret per-time-step of episode $i$.
    The fourth inequality is due to $\delta \leq 1$ and $\log T \geq 1$.
    The penultimate inequality is due to the fact that each time-step is a part of only one episode.

    What remains to be shown is that the $\max_{i \in [m]} R^{ave}_i$ term in expression \ref{eqn:minimax-exp-2}
    is $O(\delta^{\nicefrac{1}{3}} \log ^{\nicefrac{1}{3}} T)$.
    For this, we analyse $R^{ave}_i$ based on two possible cases
    depending on whether the episode $i$ had a passive phase or not, that is,
    depending on the values of $\tau_i$ and $\text{buf}$ quantities.

    \paragraph{Case $\tau_i < \text{buf}$ :} The episode $i$ has both an active and a passive phase. And we have

    \begin{align*}
        \tau_i < \frac{2}{\delta} \sqrt{\frac{\log T}{\tau_i}} \quad \iff \quad
        \tau_i < 2^{\nicefrac{2}{3}} \delta^{\nicefrac{-2}{3}} \log^{\nicefrac{1}{3}} T. \numberthis \label{eqn:minimax-tau-bound-1}
    \end{align*}

    \noindent The duration of this episode, $t_{i+1} - t_i = \text{buf} = \frac{2}{\delta} \sqrt{\frac{\log T}{\tau_i}}$. The average regret is bounded as 
    \begin{align*}
        R^{ave}_i & \leq \frac{6 \sqrt{\tau_i \log T}}{\frac{2}{\delta}\sqrt{\frac{\log T}{\tau_i}}} = 3 \delta \tau_i  \\
            & \leq 3. 2^{\nicefrac{2}{3}} \delta^{\nicefrac{1}{3}} \log^{\nicefrac{1}{3}} T & \left\{\text{From eqn. } \ref{eqn:minimax-tau-bound-1}\right\}. \numberthis \label{eqn:minimax-rave-bound-1}\\
    \end{align*}

    \paragraph{Case $\tau_i \geq \text{buf}$ :} Here, the episode $i$ does not have a passive phase. And we have
    \begin{align*}
        \tau_i \geq \frac{2}{\delta} \sqrt{\frac{\log T}{\tau_i}} \quad \iff \quad
        \tau_i \geq 2^{\nicefrac{2}{3}} \delta^{\nicefrac{-2}{3}} \log^{\nicefrac{1}{3}} T \numberthis \label{eqn:minimax-tau-bound-2}
    \end{align*}

    \noindent The duration of this episode is $t_{i+1} - t_i = \tau_i$ 
    The average regret is bounded as 

    \begin{align*}
        R^{ave}_i & \leq \frac{6 \sqrt{\tau_i \log T}}{\tau_i} = 6      \frac{\sqrt{\log T}}{\sqrt{\tau_i}} \\
            & \leq \frac{6 \log^{\nicefrac{1}{2}} T}{2^{\nicefrac{1}{3}} \delta^{\nicefrac{-1}{3}} \log^{\nicefrac{1}{6}} T} & \left\{\text{From eqn. } \ref{eqn:minimax-tau-bound-2}\right\} \\
            & = \frac{6}{2^{\nicefrac{1}{3}}} \delta^{\nicefrac{1}{3}} \log^{\nicefrac{1}{3}} T \numberthis \label{eqn:minimax-rave-bound-2}
    \end{align*}

    \noindent From equations \ref{eqn:minimax-rave-bound-1} and \ref{eqn:minimax-rave-bound-2}, we have that
    for all episodes $i$,
    $R^{ave}_i \leq O(\delta^{\nicefrac{1}{3}} \log^{\nicefrac{1}{3}} T)$.
    This completes the proof of the Theorem.
\end{proof}


\subsection{Average Regret in Episode}
\label{appn-sub-sec:ave-reg-by-epis-variables}

As an addition, in this sub-section, we show
the interplay between average regret in an episode, its total duration, and its active phase duration,
and finally illustrate our observations.

From Proof in \ref{proof:minimiax-upper}, we have that 

\begin{align*}
    R^{ave}_i = 
    \begin{dcases}
        O\left(\tau_i \delta\right) \quad & \text{if  } \tau_i < c_3 \delta^{\nicefrac{-2}{3}} \log^{\nicefrac{1}{3}} T \\
        O\left(\sqrt{\frac{\log T}{\tau_i}} \right) \quad & \text{if  } \tau_i \geq c_3 \delta^{\nicefrac{-2}{3}} \log^{\nicefrac{1}{3}} T 
    \end{dcases}
    \numberthis \label{eqn:ave-reg-by-tau}
\end{align*}

If $\tau_i \geq c_3 \delta^{\nicefrac{-2}{3}} \log^{\nicefrac{1}{3}} T$, episode $i$ has no passive phase, 
and the duration of the episode (denoted by a short-hand $d$) is $d = t_{i+1} - t_i = \tau_i$.
Then, from the second expression in \ref{eqn:ave-reg-by-tau}, we have $R^{ave}_i = O\left(\sqrt{\frac{\log T}{d}} \right)$.

Otherwise, we have $\tau_i < c_3 \delta^{\nicefrac{-2}{3}} \log^{\nicefrac{1}{3}} T$, there is a passive phase taking the duration of the episode to $d = t_{i+1} - t_i = O\left(\frac{1}{\delta}\sqrt{\frac{\log T}{\tau_i}} \right)$ which gives us $\tau_i = O\left( \frac{\log T}{\delta^2 d^2} \right)$.
We get, from the first expression in \ref{eqn:ave-reg-by-tau}, the average regret $R^{ave}_i = O(\tau_i \delta) = O\left( \frac{\log T}{\delta d^2} \right)$.

Putting the two together,

\begin{align*}
    R^{ave}_i = 
    \begin{dcases}
        O\left(\frac{\log T}{d^2 \delta} \right) \quad \quad & \text{if  } \tau_i < c \delta^{\nicefrac{-2}{3}} \log^{\nicefrac{1}{3}} T \\
        O\left(\sqrt{\frac{\log T}{d}} \right) \quad & \text{if  } \tau_i \geq c \delta^{\nicefrac{-2}{3}} \log^{\nicefrac{1}{3}} T
    \end{dcases}
    \numberthis \label{eqn:ave-reg-by-epis-dur}
\end{align*}

\begin{figure*}[!h]
    \centering
    \begin{subfigure}[b]{0.45\textwidth}
            \centering
            \includegraphics[width=\linewidth]{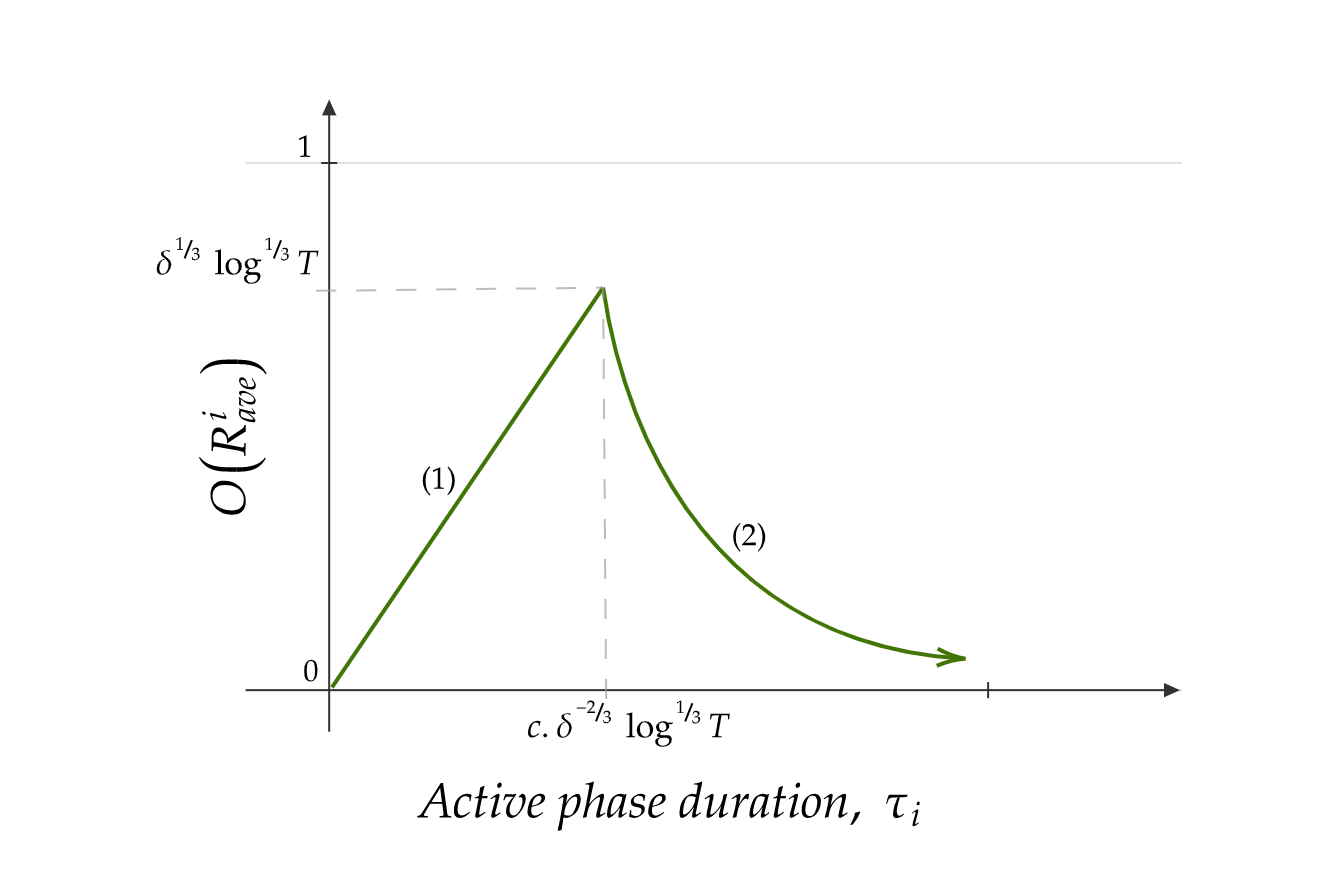}
            \caption{}
            \label{fig:ave-reg-by-tau}
    \end{subfigure}%
    \begin{subfigure}[b]{0.5\textwidth}
            \centering
            \includegraphics[width=\linewidth]{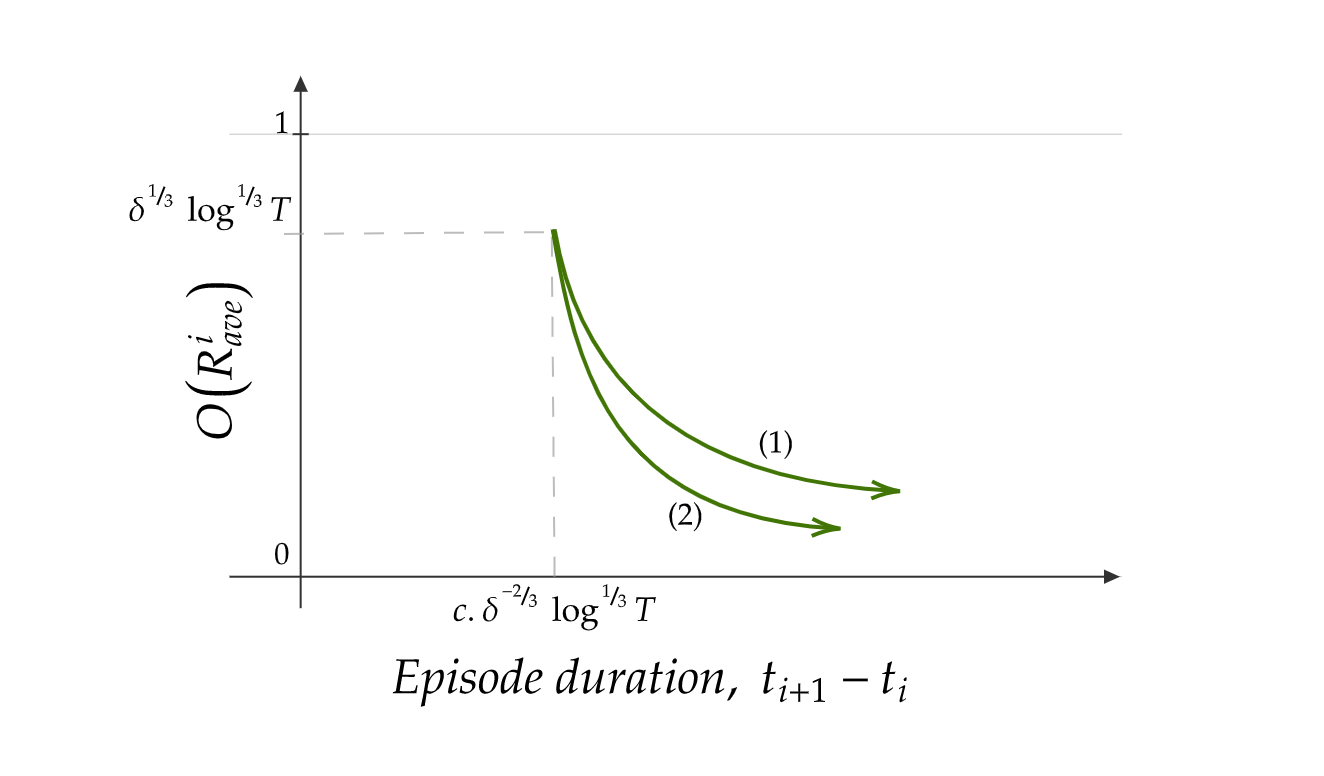}
            \caption{}
            \label{fig:ave-reg-by-epis-dur}
    \end{subfigure}
    \caption{A plot of the per-time-step regret bound in episode $i$ as a function of episode specific values;
    in fig \ref{fig:ave-reg-by-tau}, as a function of active phase duration $\tau_i$ (with curves (1), (2) as in eqn. \ref{eqn:ave-reg-by-tau}),
    and in fig \ref{fig:ave-reg-by-epis-dur} as a function of episode duration $t_{i+1}-t_i$ (with curves (1), (2) as in eqn. 
    \ref{eqn:ave-reg-by-epis-dur}) }
    \label{fig:ave-reg-plot}
\end{figure*}

We hand-curate a plot that represents these bounds in fig. \ref{fig:ave-reg-plot}.
In addition to imparting a pictorial understanding of the minimax regret behaviour, 
it also shows that the shortest possible episodes correspond to the largest average regret bounds.

\section{Proof of Theorem \ref{thm:lower-bound}}
\label{appn-sec:low-bound}

\TheoremLowerBound*

\begin{proof}

Towards proving this theorem,
first, we shall state and prove some useful information theoretic lemmas.
Then, we will divide the time horizon into smaller blocks and 
lower bound the expected regret of each block using those lemmas. 
Finally, we shall aggregate the regrets of the individual blocks by adhering to problem specific limitations,
specifically, the drift limit $\delta$, 
to arrive at the final overall lower bound.

The Change of Measure Inequality presented in Lemma \ref{lem:change-of-measure} generalizes that of \cite{garivier2016optimal}
by accommodating non-stationary reward distributions of arms.
We present the general $k$-arm case; however, we shall instantiate it with $k=2$ to suit our requirement.

\begin{lemma}[Non-stationary Change of Measure Inequality]
    \label{lem:change-of-measure}
    Let $\nu$ and $\nu'$ be two non-stationary bandit instances 
    (sets of reward distributions for each time-step)
    with $k$ arms over time horizon $[T]$. 
    For any bandit algorithm $\Alg$, for any random variable $Z$ with values in $[0,1]$ that is fully determinable from the trajectory (history) of an algorithmic run,
    $H_T$, i.e., $Z$ is $\sigma(H_T)$- measurable, we have

    \begin{equation*}
        \sum_{i=1}^{k} \sum_{t=1}^{T} \kl{\nu_{i,t}}{\nu'_{i,t}}  
        \expund{\nu}{\indicator{\Alg(t) = i} } 
        \geq
        \kl{\bern{\expund{\nu}{Z}}}{\bern{\expund{\nu'}{Z}}}
        \numberthis \label{eqn:change-of-measure}
    \end{equation*}

    where, 
    $\expund{\nu}{X}$ is the expected value of random variable $X$ under bandit instance $\nu$,
    $\Alg(t)$ is the arm played by $\Alg$ at time-step $t$,
    $KL(a,b)$ is the Kullback-Leibler divergence between distributions $a$ and $b$,
    $Ber(a)$ is the Bernoulli distribution with expectation $a$.
\end{lemma}
\begin{proof}
    We prove this inequality by establishing two intermediate results: 
    \begin{align*}
        \sum_{i=1}^{k} \sum_{t=1}^{T} \kl{\nu_{i,t}}{\nu'_{i,t}} \expund{\nu}{\indicator{\Alg(t) = i}} = \kl{\probdistunder{\nu}^{H_{T+1}}}{\probdistunder{\nu'}^{H_{T+1}}}, \quad \text{and} 
        \numberthis \label{eqn:change-of-measure-part-1} \\
        \kl{\probdistunder{\nu}^{H_{T+1}}}{\probdistunder{\nu'}^{H_{T+1}}} \geq \kl{\bern{\expund{\nu}{Z}}}{
        \bern{\expund{\nu'}{Z}}}.
        \numberthis \label{eqn:change-of-measure-part-2}
    \end{align*}
    Here,  
    $\probdistunder{\nu}^{H_{T+1}}$ (resp. $\probdistunder{\nu'}^{H_{T+1}}$) is the probability distribution under instance $\nu$ (resp. $\nu'$)
    of the algorithmic trajectory $H_{T+1}=(U_1, I_1, Y_1, \dots, U_T, I_T, Y_T, U_{T+1})$. And, $U_t, I_t, Y_t$ are random variables that correspond to internal randomness, arm pulled, and reward obtained respectively at time $t$. 

    We start with the right-hand-side (RHS) of step 1 (equation \ref{eqn:change-of-measure-part-1}) and show it's equality to the left-hand-side (LHS). By definition,

    \begin{align*}
        \kl{\probdistunder{\nu}^{H_T}}{\probdistunder{\nu'}^{H_T} } &=
        \sum_{h_{T+1}} \probunder{\nu}{H_{T+1} = h_{T+1}} \log \frac{\probunder{\nu}{H_{T+1} = h_{T+1}}}{\probunder{\nu'}{H_{T+1} = h_{T+1}}},
    \end{align*}

    where $h_{T+1} := (u_1, i_1, y_1, \dots, u_T, i_T, y_T, u_{T+1})$ is a realisation of a trajectory of a bandit algorithm. We continue by writing the sought-after divergence as
    \begin{align*}
         &\sum_{h_{T+1}} \probunder{\nu}{H_{T+1} = h_{T+1}} \log \left(\frac{f_u(u_1). \probunder{\nu}{I_1 = i_1 \vert U_1 = u_1}. \probunder{\nu}{Y_1 = y_1 \vert U_1 = u_1, I_1 =i_1}. f_u(u_2). \dots}{f_u(u_1). \probunder{\nu'}{I_1 = i_1 \vert U_1 = u_1}. \probunder{\nu'}{Y_1 = y_1 \vert U_1 = u_1, I_1 =i_1}. f_u(u_2). \dots} \right).
    \end{align*}

    The internal randomness function $f_u(\cdot)$ of the algorithm $\Alg$ is instance-agnostic. 
    Also, the probability distribution of $I_t$ under instances $\nu$ and $\nu'$ 
    are identical when conditioned upon the the trajectory 
    $(u_1, i_1, y_1, \dots, u_{t-1}, i_{t-1}, y_{t-1}, u_{t})$.
    We continue by writing the sought-after divergence as
    \begin{align*}
        & \sum_{h_{T+1}} \probunder{\nu}{H_{T+1} = h_{T+1}} \log \prod_{t=1}^{T} \frac{\nu_{i_t,t}(y_t)}{\nu'_{i_t,t}(y_t)} \\
        &= \sum_{h_{T+1}} \probunder{\nu}{H_{T+1} = h_{T+1}} \sum_{t=1}^{T} \log \frac{\nu_{i_t,t}(y_t)}{\nu'_{i_t,t}(y_t)}.
    \end{align*}

    We represent the above expression as an expectation over all possible trajectories $h_{t}$ under instance $\nu$.
    At time step $t$, the deterministic arm played $i_t$
    is replaced by the random variable $I_t$.
    We continue the derivation as follows.

    \begin{align*}
        \kl{\probdistunder{\nu}^{H_T}}{\probdistunder{\nu'}^{H_T} } &= \expund{\nu}{\sum_{t=1}^{T} \log \frac{\nu_{I_t,t}(y_t)}{\nu'_{I_t,t}(y_t)}} \\
        &= \expund{\nu}{\sum_{t=1}^{T} \log \frac{\nu_{I_t,t}(y_t)}{\nu'_{I_t,t}(y_t)} \sum_{i=1}^{k} \indicator{I_t=i} } \\
        &= \sum_{i=1}^{k} \sum_{t=1}^{T} \expund{\nu}{\indicator{I_t=i} \log \frac{\nu_{i,t}(y_t)}{\nu'_{i,t}(y_t)}} \\
        &= \sum_{i=1}^{k} \sum_{t=1}^{T} \expund{\nu}{\indicator{I_t=i} \expund{\nu}{\log \frac{\nu_{i,t}(y_t)}{\nu'_{i,t}(y_t)}} \vert \indicator{I_t=i}} \\
        &= \sum_{i=1}^{k} \sum_{t=1}^{T} \expund{\nu}{\indicator{I_t=i} KL(\nu_{i,t} , \nu'_{i,t})} \\
        &= \sum_{i=1}^{k} \sum_{t=1}^{T} KL(\nu_{i,t} , \nu'_{i,t}) \expund{\nu}{\indicator{I_t=i}}.
    \end{align*}

    The second equality is due to the fact that at every time $t$, exactly one arm is played. In the fourth equality, the inner expectation is over all realisations of $y_t$.
    This completes the proof of equation \ref{eqn:change-of-measure-part-1}.

	For the proof for inequality \ref{eqn:change-of-measure-part-2}, to avoid repetition, we refer the reader to \cite{garivier2019explore}.

	This completes the proof of the lemma.
\end{proof}

Next, we shall divide the time horizon into smaller blocks, and use this lemma to prove a lower bound on each of them individually. We shall finally aggregate them to get the final lower bound.

Divide the time horizon into blocks of size $m$, to be determined later.
We get $\nicefrac{T}{m}$\footnote{For technical clarity, we assume integrality of all quantities suitably.} blocks in total.

Throughout this section, we trade the notion of mean rewards $\mu$ (or $\mu_{i,t}$) in favour of the more generic notion of reward distributions $\nu$ (or $\nu_{i,t}$). Also, the time horizon of the block is $[m]$ instead of the global time horizon $[T]$.

Consider two reward distributions $\nu, \nu'$.
Let $\nu$ be a stationary instance with identical arms
with $\bern{\nicefrac{1}{2}}$ rewards, i.e.,
for $i \in \{1,2\}$, all time steps $t \in [m]$, 
we have $\nu_{i,t} \sim \bern{\nicefrac{1}{2}}$.

One of the two arms is played at most in half the number of time steps, 
i.e., $\exists i \in \{1,2\} : \expund{\nu}{N_i} \leq \nicefrac{m}{2}$, 
where $N_i$ is the number of times arm $i$ is played in the block (by the end of time step $m$).
Without loss of generality, assume that it is arm $i=1$ that satisfies the above condition, 
i.e., $\expund{\nu}{N_1} \leq \nicefrac{m}{2}$.

We construct bandit problem instance $\nu'$ in such a way that the lesser played arm 
in instance $\nu$ (arm $1$) is optimal in $\nu'$.

In $\nu'$, arm $2$ has a stationary (for all times $t \in [m]$) $\bern{\nicefrac{1}{2}}$ reward distribution. 
Whereas, arm $1$ has a $\bern{\nicefrac{1}{2}}$ reward distribution 
at the beginning and ending time-step of the block, but,
has a reward distribution with larger means (than $\nicefrac{1}{2}$)
in the intervening time-steps.
Precisely,
\begin{equation*}
\numberthis
\label{eqn:lb-bad-instance}
    \nu'_{i,t} =
    \begin{dcases}
        \bern{\dfrac{1}{2}} \quad  & \text{if  } i=2\\
        \bern{\dfrac{1}{2} + \dfrac{t-1}{m}.\varepsilon} \quad & \text{if  } i =1, t \leq \ceil{\dfrac{m}{2}}\\
        \bern{\dfrac{1}{2} + \dfrac{m-t}{m}.\varepsilon} \quad & \text{if  } i =1, t > \ceil{\dfrac{m}{2}}.
    \end{dcases} 
\end{equation*}

We apply the non-stationary change of measure inequality stated in Lemma \ref{lem:change-of-measure} to the instances $\nu$ and $\nu'$
with a choice of $Z=\nicefrac{N_1}{m}$, 
the play fraction of the arm that is underplayed in $\nu$, 
but is optimal in $\nu'$. 

We first upper bound the LHS before plugging it into the inequality:
\begin{align*}
     &\sum_{i=1}^{k} \sum_{t=1}^{m} \kl{\nu_{i,t}}{\nu'_{i,t}}
        \expund{\nu}{\indicator{\Alg(t) = i} } \\
    =& \sum_{t=1}^{m} \kl{\nu_{1,t}}{\nu'_{1,t}} 
    \expund{\nu}{\indicator{\Alg(t) = 1} }  \\
    =& 
    \sum_{t=1}^{\ceil{\frac{m}{2}}} \kl{\bern{\frac{1}{2}}}{\bern{\frac{1}{2} + \frac{t-1}{m}.\varepsilon}}
    \expund{\nu}{\indicator{\Alg(t) = 1} } 
    + \\ 
    & \quad \sum_{t=\ceil{\frac{m}{2}}+1}^{m} \kl{\bern{\frac{1}{2}}}{\bern{\frac{1}{2} + \frac{m-t}{m}.\varepsilon}} 
    \expund{\nu}{\indicator{\Alg(t) = 1} } \\
    \leq & \sum_{t=1}^{\ceil{\frac{m}{2}}} \kl{\bern{\frac{1}{2}}}{\bern{\frac{1+\varepsilon}{2}}} 
    \expund{\nu}{\indicator{\Alg(t) = 1} } 
    + \\ 
    & \quad \sum_{t=\ceil{\frac{m}{2}}+1}^{m} \kl{\bern{\frac{1}{2}}}{\bern{\frac{1+\varepsilon}{2}}} 
    \expund{\nu}{\indicator{\Alg(t) = 1} } 
    \numberthis \label{eqn:low-bou-strengthen} \\
    =& \kl{\bern{\frac{1}{2}}}{\bern{\frac{1+\varepsilon}{2}}} \sum_{t=1}^{m}  
    \expund{\nu}{\indicator{\Alg(t) = 1} }
    \leq \varepsilon^2. \expund{\nu}{N_1}. \numberthis \label{eqn:lower-bound-lhs-up-bou}
\end{align*}

The first equality is due to the fact that arm $2$ is identical in both the instances, i.e.,
$\forall t \in [m]$, we have $\kl{\nu_{2,t}}{\nu'_{2,t}} = 0$.

The inequality \ref{eqn:low-bou-strengthen} upper bounds the information gathered about the distinguishability of the arm $1$ (from stationary arm $1$ with $\bern{\nicefrac{1}{2}}$ rewards in instance $\nu$) as if it were statinonary with $\bern{\frac{1+\varepsilon}{2}}$ rewards at all times $t \in [m]$ in $\nu'$.

We now plug the inequality \ref{eqn:lower-bound-lhs-up-bou} into inequality \ref{eqn:change-of-measure} with a choice of $Z = \nicefrac{N_1}{m}$ to get
\begin{align*}
    \varepsilon^2 \expund{\nu}{N_1} &\geq \kl{\bern{\frac{\expund{\nu}{N_1}}{m}}}{
    \bern{\frac{\expund{\nu'}{N_1}}{m}}} \\ 
    & \geq 2 \left( \frac{\expund{\nu}{N_1}}{m} - \frac{\expund{\nu'}{N_1}}{m}  \right)^2 \tag{By Pinsker's Inequality} \\
    \implies \sqrt{\frac{1}{2}. \varepsilon^2 \expund{\nu}{N_1}} 
    & \geq \frac{\expund{\nu'}{N_1}}{m} - \frac{\expund{\nu}{N_1}}{m} \\
    \frac{\expund{\nu'}{N_1}}{m} & \leq \frac{\expund{\nu}{N_1}}{m} + \sqrt{\frac{1}{2}. \varepsilon^2 \expund{\nu}{N_1}} \\
    & \leq \frac{1}{2} + \sqrt{\frac{1}{2}. \varepsilon^2. \frac{m}{2}}.
\end{align*}

The final inequality is due to $\expund{\nu}{N_1} \leq \nicefrac{m}{2}$. Fixing $\varepsilon = \sqrt{\nicefrac{1}{4m}}$, 
we get $\frac{\expund{\nu'}{N_1}}{m} \leq \nicefrac{3}{4}$. Thus, the sub-optimal arm $2$ in instance $\nu'$ is played 
for more than a constant fraction of times in expectation,
i.e., $\expund{\nu'}{N_2} \geq \nicefrac{m}{4}$. In instance $\nu'$, let this sub-optimal arm $2$ be played at time steps $t_1, t_2, \dots, t_x$ for some $x \geq \nicefrac{m}{4}$.
Then, the expected regret in the block (denoted by $R^b(\Alg)$) is 

\begin{align*}
    \expund{\nu'}{R^b(\Alg)} = \sum_{i=1}^x \mu_{1, t_x} - \mu_{2,t_x} \\
\end{align*}

By design (in equation \ref{eqn:lb-bad-instance}), the set of gaps, $\left\{\abs{\mu_{1,t} - \mu_{2,t}}\right\}_{t \in [m]}$, between the arms throughout the block $[m]$ is
$\{ 0,0,\frac{\varepsilon}{m},\frac{\varepsilon}{m},\frac{2\varepsilon}{m}, \dots, \frac{\varepsilon}{2} \}$.

We lower bound $\sum_{i=1}^x \mu_{1, t_x} - \mu_{2,t_x}$ 
with the least possible sum of $x$ values from the set of gaps:
\begin{align*}
    \expund{\nu'}{R^b(\Alg)} &= \sum_{i=1}^x \mu_{1, t_x} - \mu_{2,t_x} \\
    & \geq \sum_{i=1}^{m/8} 2. \frac{(i-1). \varepsilon}{m} \\
    & = \frac{2 \varepsilon}{m} \sum_{i=0}^{m/8 - 1} i \\
    & = \frac{2 \varepsilon}{m}. \frac{m^2 - 8m}{64} = \frac{1}{64}. \frac{m^2 - 8m}{m^{3/2}} \\
    & = \frac{1}{64}. \left( m^{1/2} - 8/m^{1/2} \right) = \Omega\left(\sqrt{m}\right). \numberthis \label{eqn:lb-block-lb}
\end{align*}

The penultimate line is due to $\varepsilon = \sqrt{\nicefrac{1}{4m}}$.

Also, at each time step in $t \in [m]$, the gap is upper bounded as $\abs{\mu_{1,t} - \mu_{2,t}} \leq \varepsilon$.
So, the regret in the block is trivially upper bounded as follows:
\begin{align*}
    \expund{\nu'}{R^b(\Alg)} & \leq m . \varepsilon = m. \sqrt{\nicefrac{1}{4m}} = O\left(\sqrt{m} \right). \numberthis \label{eqn:lb-block-ub}
\end{align*}

The following lemma is a conseqence of 
\ref{eqn:lb-block-lb} and \ref{eqn:lb-block-ub}.

\begin{lemma}
\label{lem:block-regret-lower-bound}
    For a block of time period $m$,
    there exists a bandit instance $\nu'$ (as in expression \ref{eqn:lb-bad-instance}) such that,
    for any algorithm $\Alg$, its expected regret $\expund{\nu'}{R^b(\Alg)}$ is $\Theta(\sqrt{m})$.
\end{lemma}

Next, we aggregate the regret across $\nicefrac{T}{m}$ blocks to get the overall regret.
Note that the instances $\nu$ and $\nu'$ have a drift limit implication as follows:
\begin{align*}
    \delta \geq \frac{\varepsilon}{m} \quad \iff \quad
    \delta \geq \frac{1}{2m^{\nicefrac{3}{2}}} \quad \iff \quad
    m^{\nicefrac{3}{2}} \geq \frac{1}{2 \delta} \quad \iff \quad
    \sqrt{m} \geq (2\delta)^{\nicefrac{-1}{3}}. \numberthis \label{eqn:lb-m-delta-relation}
\end{align*}



Now, the total regret is lower bounded by the number of blocks multiplied by the lower bound of regret within each block; thus,
\begin{align*}
\expectation{R(\Alg)} &= \frac{T}{m} \times \Theta(\sqrt{m}) \\ 
&= \Omega \left( T \delta^{1/3} \right). \tag{From \ref{eqn:lb-m-delta-relation}}
\end{align*}

This completes the proof of the theorem.
\end{proof}

\section{Experiments}
\label{appn-sec:experiments}

\subsection{Modified Algorithm}
\label{appn-subsec:modified-algo}

In this subsection, we mention a modification of our $\AlgGE$ algorithm as discussed in Remark \ref{rem:empirical-tweaks}.
We tweak the original algorithm (Algorithm \ref{alg:gauge-enjoy}) in two places 
to obtain the modified algorithm (Algorithm \ref{alg:gauge-enjoy-modified}).

First, in Line \ref{m-line:compute-buffer}, we use a tighter (larger) sub-optimality buffer $\text{buf} = \frac{\lambTil}{6}$ 
as opposed to the original $\text{buf} = \frac{2}{\delta} \sqrt{\frac{\log T}{\tau_i}}$.
Second, the duration (towards the past) from current time $t$
in which the time-step of guaranteed true gap can lie (Claim \ref{clm:recent-gap-existance})
can be a much shorter $2w$, the size of the statistical test window, as opposed to the original $\tau_i$, the duration of the active phase.
This change is reflected in Lines \ref{m-line:stat-test-window}, \ref{m-line:check-snooze-condition}, and \ref{m-line:snooze}.

\begin{algorithm}[h]
\small
\caption{$\AlgGE$(m): A modified algorithm to play the non-stationary slowly-varying bandit problem instance}
\label{alg:gauge-enjoy-modified}
{\bf Input:} Time horizon $T$, a set $2$ arms $\Arms = \{1,2\}$ with sample access, the drift limit $\delta$.\\
{\bf Output:} Play an arm for every time-step. 
\begin{algorithmic}[1]
    \State Initialize set of active arms $A = \{1, 2\}$,
    set of snoozed arms $S = \emptyset$.
    \State Initialize episode index $i \leftarrow 1$, $\tau_1 = 0$.
    \For{$t = 1,2,\dots,T$}
        \State $x \leftarrow$ Least recently pulled arm in $A$.
        \Comment{Play active arms in round-robin fashion}
        \State Pull arm $x$ and observe reward $\muhat_{x,t}$.
        \If{$\exists$ arms $a,b \in A$, $\exists \lambTil \in [0,1]$ s.t. $a \lambTilGreat b$}
        \Comment{As in Definition \ref{def:lambda-better}}
            \State Statistical test success time, $g_i = t$
            \State Active phase duration, $\tau_i = g_i - t_i$.
            \State Statistical test window, $w := \ceil{\dfrac{c_1 \log T}{\lambTil^2}}$.
            \label{m-line:stat-test-window}
            \State Sub-optimality buffer, $\text{buf} = \dfrac{\lambTil}{6}$
            \label{m-line:compute-buffer}
            \If{$\text{buf} > 2w$}
            \label{m-line:check-snooze-condition}
                \State $A \leftarrow A \setminus \{b\}$
                \State $S \leftarrow S \cup \{(b, g_i - 2w + \text{buf})\}$
                \label{m-line:snooze}
                \Comment{Snooze arm}
            \Else
                \State $i \leftarrow i+1, t_i \leftarrow t$
                \Comment{End episode without passive period}
            \EndIf
        \EndIf
 
        \If{$\exists (x,s) \in S: s \geq t$}
            \State $S \leftarrow S \setminus \{(x,s)\}, 
                     A \leftarrow A \cup \{x\}$
            \Comment{Respawn arm}
            \State $i \leftarrow i+1, t_i \leftarrow t$
            \Comment{End episode for passive period elapses}
        \EndIf
    \EndFor
\end{algorithmic}
\end{algorithm}

\subsection{Missing Figures from Section \ref{sec:algorithm} and Section \ref{sec:experiments}}
\label{appn-subsec:missing-figures}

\begin{figure*}[!h]
    \centering
    \begin{subfigure}[b]{0.48\textwidth}
            \centering
			\includegraphics[width=\linewidth]{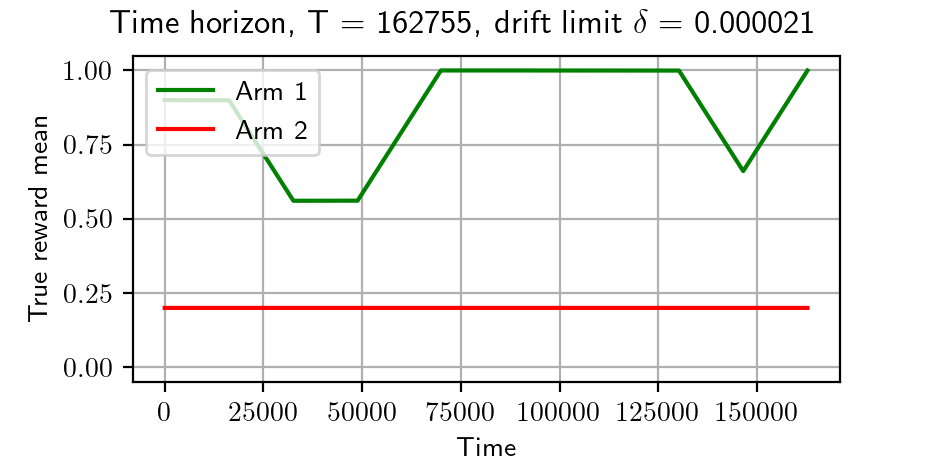}
            \caption{}
            \label{fig:instance-low-delt-well-sep}
    \end{subfigure}
    \begin{subfigure}[b]{0.48\textwidth}
            \centering
		    \includegraphics[width=\linewidth]{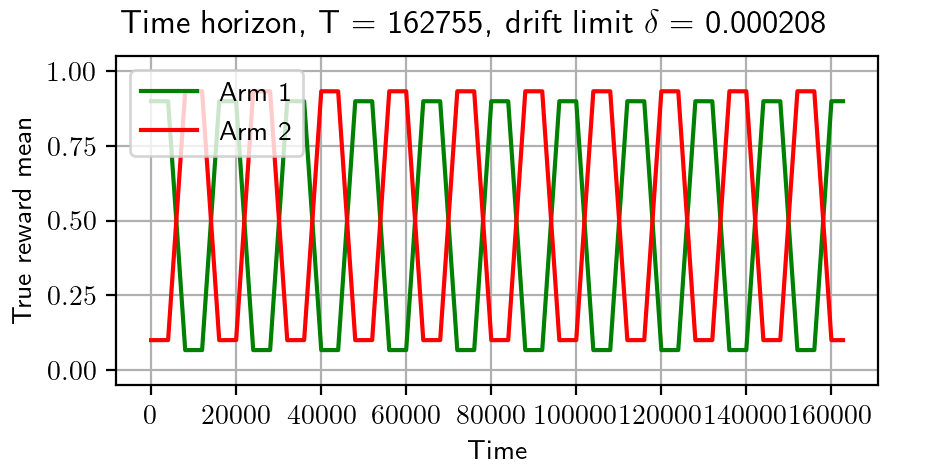}
            \caption{}
            \label{fig:instance-high-delt-oscillating-arms}
    \end{subfigure}%
    \caption{Illustration of problem instances, i.e., the true reward means of the arms. 
    In (a), the arms are well separated with no change in optimal arm's identity throughout. Here, the drift limit $\delta=\nicefrac{1}{10. c_9. \log T} \simeq 0.000021$.
    In (b), the arms experience short stretches of stationarity and drift alternately. The optimal arm's identity toggles with every drift. Here, we have a relatively large drift limit $\delta=\nicefrac{1}{c_9. \log T} \simeq 0.00021$}
\end{figure*}

\begin{figure*}[!h]
    \centering
    \begin{subfigure}[b]{0.5\textwidth}
            \centering
        	\includegraphics[width=\linewidth]{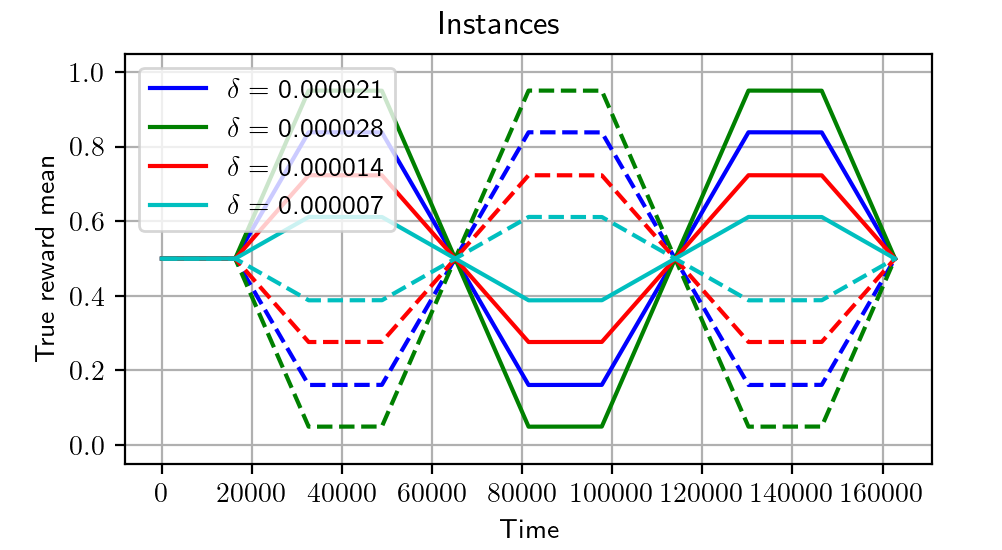}
            \caption{}
	    	\label{fig:instance-snoozeit-multi-delta-1}
    \end{subfigure}
    \begin{subfigure}[b]{0.46\textwidth}
            \centering
        	\includegraphics[width=\linewidth]{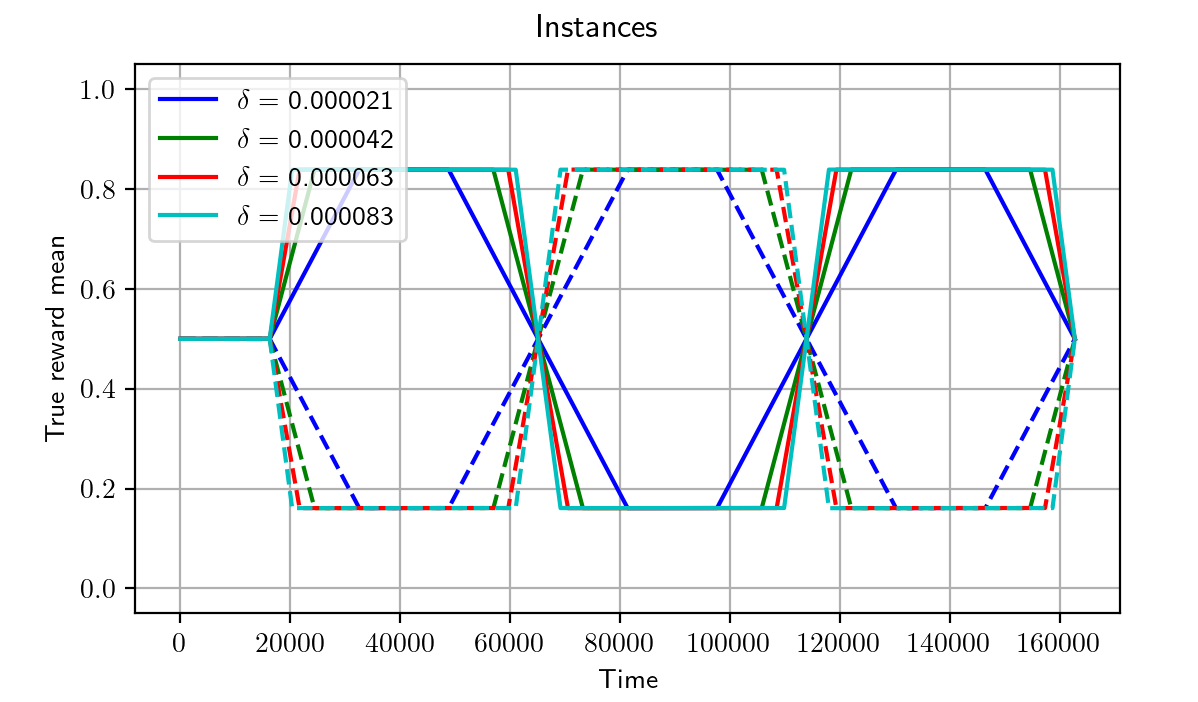}
            \caption{}
    		\label{fig:instance-snoozeit-multi-delta-2}
    \end{subfigure}%
    \caption{Two sets of 4 structurally similar problems with varying drift limits $\delta$.
    In (a), the instances have common periods of stationarity and drift. The amount of drift varies with the corresponding $\delta=a, 2a, 3a, 4a$, for $a=\nicefrac{1}{c_{10}\log T} \simeq 0.000007$ values.
    In (b), the instances have equal total cumulative drift. To achieve that drift, the duration of drift varies with the corresponding $\delta=a, 2a, 3a, 4a$, for $a=\nicefrac{1}{c_{11}\log T} \simeq 0.000021$ values.}
\end{figure*}

\begin{figure*}[!h]
    \centering
    \includegraphics[width=\linewidth]{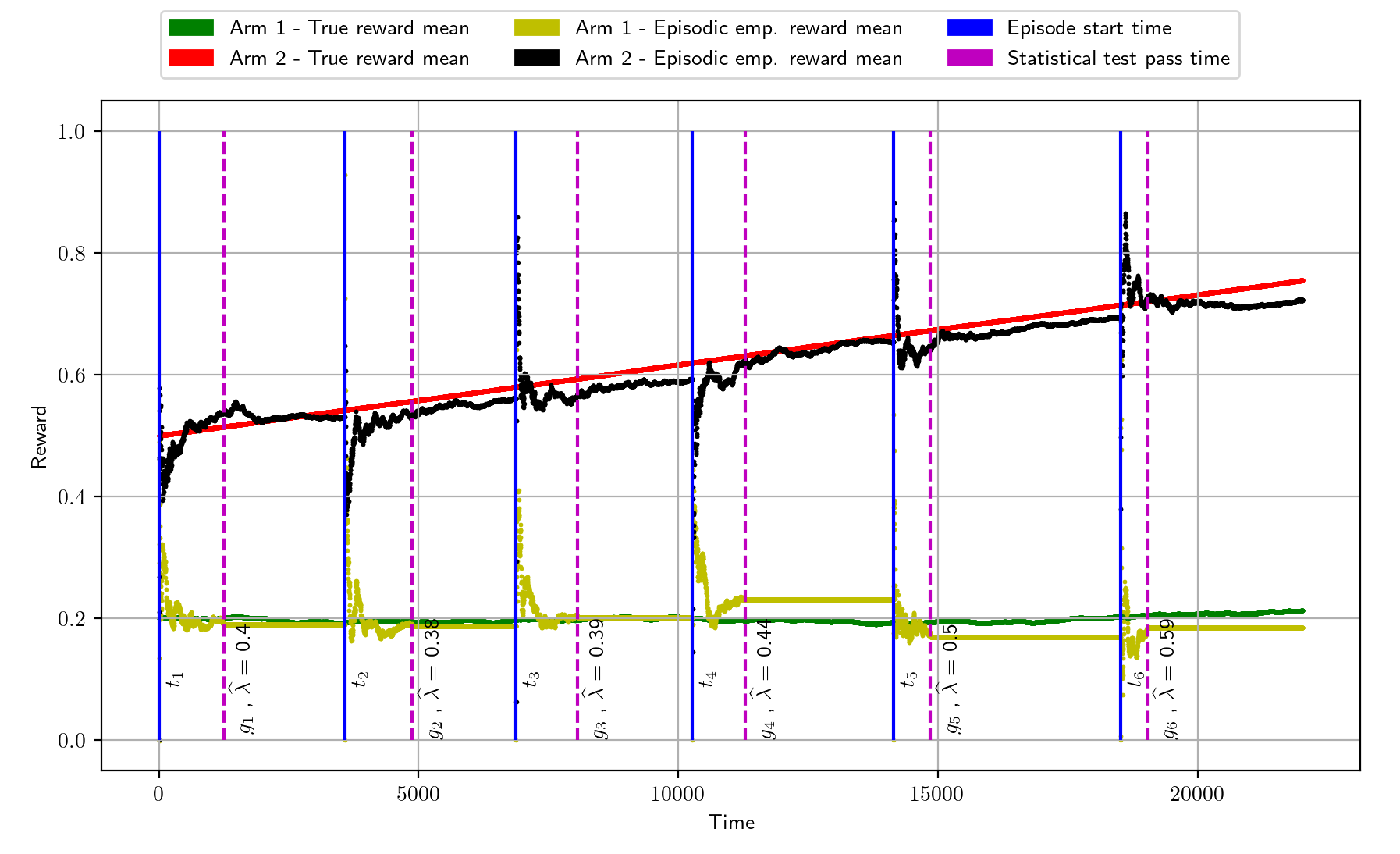}
    \caption{Illustration of the trajectory of $\AlgGE$ for a problem instance whose gap, $\Delta_t$, and thus the detectable gap, $\lambda_t$ (not shown in picture) increases with time. One can observe that the statistical test passes sooner (a shorter active phase $[t_i+1,g_i]$ of episode $i$) for larger observed gaps, $\lambTil$s. Also, the snooze period is more (a longer passive phase $[g_i+1, t_{i+1}]$) for larger observed gaps.
    The empirical means plotted are measured from the beginning of the episode till the current time-step. Note that the empirical means of the sub-optimal arm remains unchanged during the passive phase due to it not being played.}
    \label{fig:alg-trajectory-1}
\end{figure*}

\begin{figure*}[!h]
    \centering
    \includegraphics[width=\linewidth]{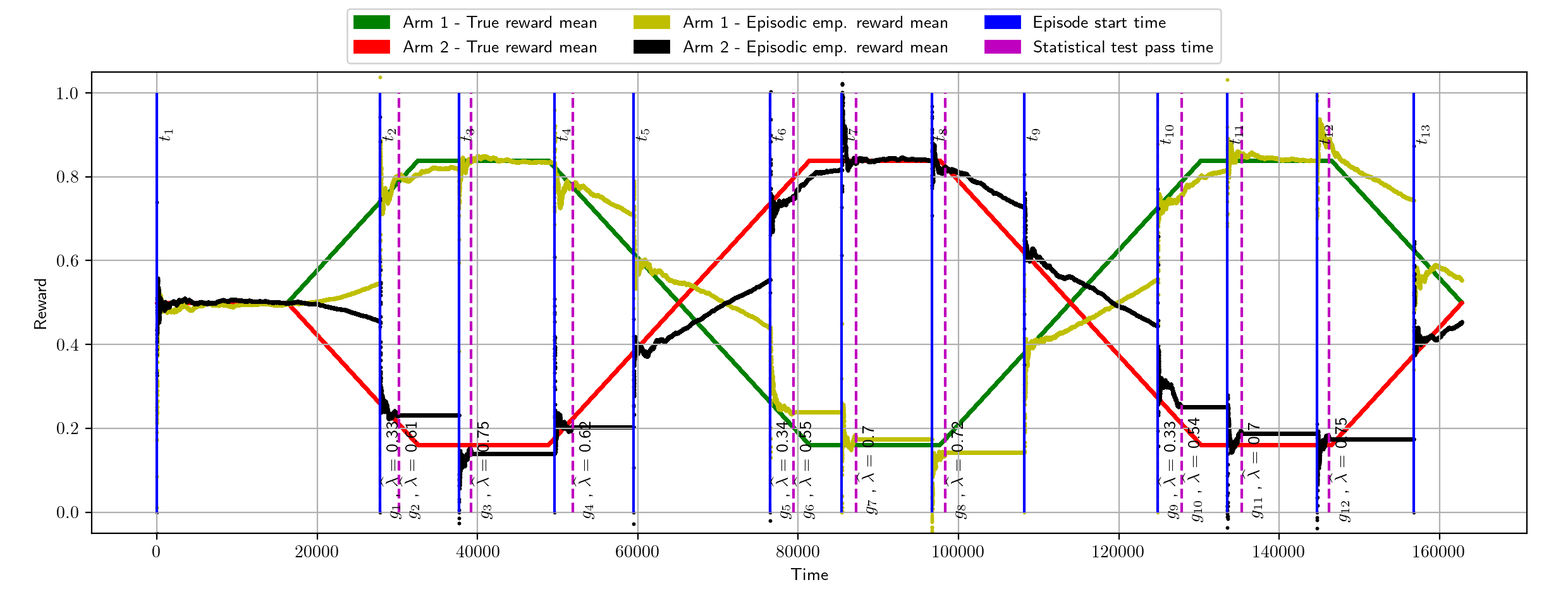}
    \caption{Illustration of the trajectory of $\AlgGE$ for a problem instance with oscillating arms.
    Note that episodes $1, 5,$ and $9$ do not have a passive phase. Essentially, compared with the length of the active phases, the gaps detected $\lambTil$s were not sufficiently large to warrant snoozing the sub-optimal arm.
    However, stretches of time where the arms are well separated enjoys passive phases (of no regret).}
    \label{fig:alg-trajectory-2}
\end{figure*}
\vfill

\end{document}